\documentclass{article}
\usepackage[T1]{fontenc}
\usepackage{iclr2027_conference,times}
\usepackage{scalecast_preprint}
\usepackage{amsmath,amssymb,amsfonts,graphicx,booktabs,tabularx,multirow,threeparttable,colortbl}
\usepackage{xcolor}
\usepackage{hyperref,url}
\usepackage{placeins}
\usepackage{enumitem}
\usepackage{graphicx}
\usepackage{wrapfig}
\newcommand{\model}{ScaleCast}
\definecolor{purple}{HTML}{763E9B}

\hypersetup{colorlinks=true,linkcolor=blue,citecolor=blue,urlcolor=blue}

\title{Learning Kilometer-Scale Weather Prediction with Global-Regional Alignment}
\author{%
\begin{minipage}{\dimexpr\textwidth-2\tabcolsep\relax}
\centering\normalfont\small
Guowen Li$^{1,*}$, Yang Liu$^{2,*}$, Yujie Wang$^{3}$, Qiuyan Sun$^{1}$,\\
Haoyuan Liang$^{1}$, Juepeng Zheng$^{3,4,\dagger}$, Hong Cheng$^{2}$, Haohuan Fu$^{1,4}$\\[4pt]
{\footnotesize
$^{1}$Tsinghua University \quad $^{2}$The Chinese University of Hong Kong\\
$^{3}$Sun Yat-Sen University \quad $^{4}$NSCC-SZ\\[2pt]
\texttt{zhengjp8@mail.sysu.edu.cn}\\
$^{*}$Equal contribution. \quad $^{\dagger}$Corresponding author.}
\end{minipage}%
}
\date{}
\hypersetup{
  pdftitle={Learning Kilometer-Scale Weather Prediction with Global-Regional Alignment},
  pdfauthor={Guowen Li, Yang Liu, Yujie Wang, Qiuyan Sun, Haoyuan Liang, Juepeng Zheng, Hong Cheng, Haohuan Fu},
  pdfsubject={Regional weather forecasting with Global-Regional Alignment},
  pdfkeywords={ScaleCast, regional weather forecasting, global-regional alignment}
}
\begin{document}
\maketitle

\begin{abstract}
Kilometer-scale regional weather forecasting is essential for local weather warnings and weather-sensitive decisions. Existing data-driven approaches often rely on numerical forecasts for large-scale guidance or require additional training of global forecasting components. Pretrained global weather models offer an efficient source of large-scale forecasts, motivating their reuse to guide high-resolution regional prediction. However, this coupling requires aligning global and regional representations across different grids and integrating global guidance with local interactions to advance regional states.
We propose ScaleCast, a regional forecasting framework that addresses these challenges through Global--Regional Alignment. Its Global--Regional Conversion module aligns joint global and regional representations with regional locations, while the Global--Regional Alignment and Dynamics block combines aligned guidance with regional neighborhood interactions.
Experiments using ERA5 global analyses on a $0.25^\circ$ grid and CERRA regional reanalysis at 5.5 km spacing demonstrate improved regional forecasts across surface and upper-air variables, with a single trained model supporting multiple global forecast drivers (i.e., Pangu-Weather, GraphCast, and HRES) without specific retraining. Fine-tuning on HRRR at 3 km spacing further demonstrates the framework's adaptability to a different regional domain and spatial resolution. Windstorm case studies show improved cyclone positioning and core-pressure estimates, while comparisons with HadISD station observations show closer agreement with local temperature and humidity changes.

\end{abstract}


\section{Introduction}
Regional weather forecasting supports weather warnings and weather-sensitive decisions by predicting atmospheric conditions at spatial scales relevant to local impacts. For example, near-surface temperature forecasts inform assessments of local heat and cold risks, while mean sea level pressure forecasts help characterize the position and central pressure of extratropical cyclones. To generate high-resolution regional predictions, conventional limited-area numerical weather prediction (NWP) models typically use global forecasts to supply time-varying lateral boundary conditions, either directly or through a larger regional model \citep{tudor2010boundary}. These boundary conditions represent the influence of the evolving atmosphere outside the regional domain. Initialized from a regional data assimilation system, regional models numerically integrate the governing atmospheric equations on a finer grid under the boundary constraints \citep{clark2016convection,adamov2025limited}. However, this workflow requires computationally demanding regional integration and depends on the availability of global forecasts to supply boundary conditions, creating trade-offs between spatial resolution, update frequency, and forecast delivery time \citep{pathak2026stormcast}.

Recent data-driven methods offer efficient alternatives through two ways of incorporating large-scale atmospheric information. The first uses external fields for regional reconstruction or state evolution. CorrDiff learns coarse-to-fine reconstruction from global reanalysis \citep{mardani2025corrdiff}, while graph-based models such as GraphLAM, Hi-LAM, and the regional version of Graph-EFM incorporate boundary information from regional NWP data \citep{oskarsson2023graph,oskarsson2024probabilistic}. More recent approaches use global forecasts directly such as Integrated Forecasting System (IFS) forecasts \citep{adamov2025limited} or Global Forecast System (GFS) synoptic fields \citep{pathak2026stormcast}. Although these methods accelerate regional prediction, configurations driven by numerical forecasts still depend on upstream NWP outputs and their availability. The second develops global forecasting capabilities within the modeling pipeline. Bris combines global pretraining with refinement on a stretched grid \citep{nipen2026bris}, whereas OneForecast connects its global and regional predictors through neural nesting \citep{gao2025oneforecast}. These designs support global--regional coupling but incur additional training costs for their global components. 

Pretrained global weather models such as Pangu-Weather and GraphCast have demonstrated strong forecast skill and become readily available for practical forecasting applications, with substantially lower inference costs than conventional NWP systems \citep{bi2023pangu,lam2023graphcast}. These advances provide an established source of efficient large-scale forecasts on which regional prediction systems can build. However, their coarse output grids do not directly resolve kilometer-scale regional detail. This motivates us to investigate how existing global weather models can support the development of data-driven regional forecasting: using their forecast trajectories as large-scale guidance while learning to predict the evolution of high-resolution regional states.

\begin{wrapfigure}{r}{0.52\textwidth}
    \centering
    \includegraphics[width=\linewidth]{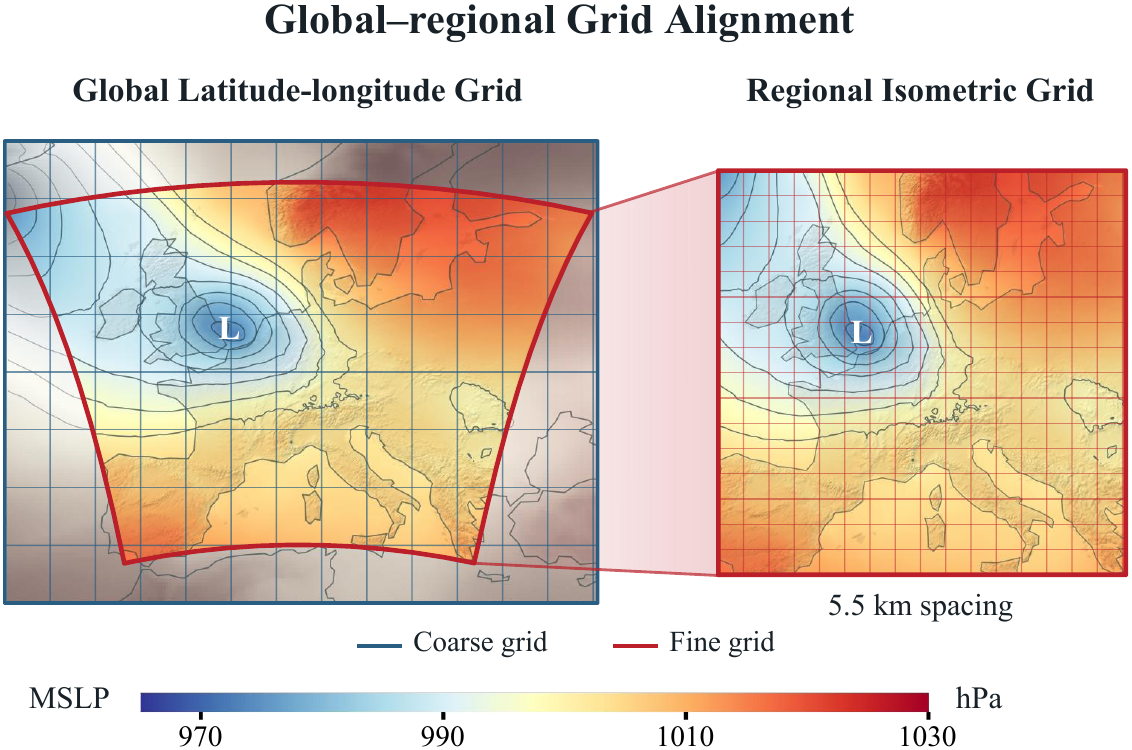}
    \caption{\textbf{Coarse global and finer regional projected grids.} The two grids sample the same atmospheric system at different locations and spatial resolutions. The surrounding gray shading denotes the boundary region. Background colors indicate mean sea level pressure (MSLP).}
    \label{fig:global-regional-alignment}
\end{wrapfigure}

However, coupling pretrained global forecasts with high-resolution regional states presents two technical challenges.
\textbf{First, global--regional grid alignment.}
Global fields commonly use latitude--longitude grids, whereas regional fields often use uniformly spaced projected grids (Figure~\ref{fig:global-regional-alignment}). Differences in sampling locations, spacing, and spatial scales complicate cross-grid fusion~\citep{mousavi2025rigno}. Direct index-wise fusion can mismatch geographic locations, while geographic remapping alone does not establish prediction-relevant feature relationships. The challenge is to align global and regional representations for regional forecasting.
\textbf{Second, regional interaction under global guidance.}
After alignment, global guidance must be integrated with regional conditions to predict local evolution. For example, the temperature response to an approaching cold front depends on both the current local state and neighboring temperature and wind conditions~\citep{xu2025yinglong,abdi2026hrrrcast}. The challenge is therefore to combine aligned global guidance, the current regional state, and its neighborhood into spatially coupled updates. Each step should advance the existing regional state, rather than reconstruct a fine-scale field from the global forecast.

We propose \model{}, a forecasting framework centered on \textit{Global--Regional Alignment}. Its Global--Regional Alignment and Dynamics (GRAD) block integrates token alignment and updates to condition regional evolution on global guidance. Within GRAD, the \textbf{G}lobal--\textbf{R}egional \textbf{Conv}ersion (GRConv) module projects joint global--regional representations onto regional token positions. Residual pathways retain the regional state, enabling autoregressive updates guided by global forecasts.
We train the framework using the European Centre for Medium-Range Weather Forecasts (ECMWF) Reanalysis v5 (\textbf{ERA5}) global analyses \citep{hersbach2020era5} and Copernicus European Regional Reanalysis (\textbf{CERRA}) \citep{ridal2024cerra}, whose native grid spacing is 5.5~km. Regional rollout experiments use Pangu-Weather, GraphCast, and the ECMWF high-resolution forecast (HRES) as global drivers and CERRA as the gridded verification reference. The results show lower errors across nine variables than the evaluated other forecasting baselines. Windstorm evaluation against the Hodges reference tracks in the \textbf{C3S European windstorm} dataset \citep{c3s2025windstorm} shows lower position errors in the examined case. Station comparisons with quality-controlled \textbf{HadISD} observations \citep{dunn2012hadisd,dunn2016expanding} show closer agreement in temperature and relative humidity. Transfer experiments on High-Resolution Rapid Refresh (\textbf{HRRR})~\citep{dowell2022hrrr} further assess cross-dataset generalization.
Our contributions are:
\begin{itemize}[leftmargin=*]
    \item We propose \model{}, a regional forecasting framework centered on Global--Regional Alignment. GRAD combines this alignment with attention and updates the regional state. Experiments on HRRR further show the framework's transferability.

    \item Regional rollout evaluation demonstrates improved surface and upper-air forecasts through $+30$~h. Reuse with Pangu-Weather, GraphCast, and HRES improves on the corresponding global forecasts without driver-specific retraining and outperform CERRA NWP on all nine variables.

    \item We validate \model{} in real-world scenarios. The windstorm case shows lower cyclone-position errors and core-pressure bias, while station comparisons show closer agreement with observed temperature and humidity changes, demonstrating benefits beyond gridded verification.
\end{itemize}

\section{Related work}
\paragraph{Global weather forecasting.}
\vspace{-5pt}
Global numerical weather prediction systems such as ECMWF-IFS integrate discretized atmospheric equations and parameterize unresolved processes, at substantial computational cost \citep{bauer2015quiet}. Learning-based alternatives approximate atmospheric evolution with learned transition operators, enabling fast inference after training. Early graph-based forecasting~\citep{keisler2022forecasting} and FourCastNet~\citep{pathak2022fourcastnet} explored this approach through message passing and Fourier operators. Earth-aware transformers in Pangu-Weather~\citep{bi2023pangu} and multiscale graphs in GraphCast~\citep{lam2023graphcast} subsequently achieved lower errors than operational IFS baselines across many evaluated variables and lead times. FengWu~\citep{chen2025operational} and FuXi~\citep{chen2023fuxi} use multivariable learning and cascaded forecast models, respectively, to extend useful medium-range prediction. Further developments include graph--transformer forecasting in AIFS \citep{lang2024aifs}, hybrid numerical--neural modeling in NeuralGCM \citep{kochkov2024neural}, and large-scale pretraining in Aurora \citep{bodnar2025aurora}. Diffusion-based GenCast extends this progress to probabilistic forecasting, outperforming ECMWF's ensemble system on most evaluated targets \citep{price2025gencast}. The demonstrated skill and low inference cost of pretrained global models motivate our use of their forecast fields as regional guidance, concentrating learning on regional prediction rather than retraining a global forecaster.

\vspace{-5pt}
\paragraph{Regional weather forecasting.}
Numerical limited-area models couple regional dynamics to larger-scale forecasts through time-varying lateral boundary conditions \citep{warner1997lateral}. Learned regional forecasting develops three partly overlapping routes. Boundary-forced models use multiscale and probabilistic graph representations \citep{oskarsson2023graph,oskarsson2024probabilistic}, separate forcing encoders with optional interior overlap \citep{adamov2025limited}, or boundary-conditioned regional networks \citep{xu2025yinglong,larsson2025diffusion}. A second route conditions regional updates on full-domain synoptic information, as in StormCast and HRRRCast \citep{pathak2026stormcast,abdi2026hrrrcast}; STCast further learns spatial alignment through attention with a geographic prior \citep{chen2026stcast}. A third route organizes global--regional exchange through neural nesting in OneForecast \citep{gao2025oneforecast} or a globally connected, regionally refined grid in Bris \citep{nipen2026bris}. The approaches differ in where external information enters and how the two spatial scales interact. \model{} uses GRConv to project concatenated global and regional tokens onto regional positions. GRAD then models input-dependent neighborhood interactions before a decoder predicts a regional state increment. In our experiments, our method achieves better performance than the baselines.

\vspace{-5pt}
\paragraph{Regional post-processing and downscaling.}
Post-processing corrects forecast biases or calibrates predictive distributions against observations \citep{schulz2022machine,hohlein2024postprocessing}, whereas spatial downscaling learns coarse-to-fine relationships, including CNN-based climate downscaling \citep{banomedina2022deepesd}, transformer-based reanalysis reconstruction \citep{perez2024transformer}, and geography-aware super-resolution \citep{xu2026geofar}. Generative approaches model conditional fine-scale variability through adversarial learning \citep{harris2022stochastic} or diffusion, including ERA5--CERRA wind downscaling and CorrDiff's regression--residual decomposition \citep{merizzi2024wind,mardani2025corrdiff}. Temporal dependence and driver reuse are also studied: EnScale-t conditions on a preceding high-resolution state \citep{schillinger2026enscale}, while universal diffusion downscaling transfers a reanalysis-trained model across upstream forecasts \citep{molinaro2026universal}. Our distinction therefore lies in the prediction operator, not simply the use of forecast inputs or temporal context.
\model{} learns cross-grid feature coupling to predict an increment of the existing regional state, rather than a correction to an upstream forecast or a same-time coarse-to-fine reconstruction. At inference, \model{} uses forecasts from pretrained global weather models as boundary guidance for regional forecasting.


\section{Methodology}
\label{sec:method}

\begin{figure}[t]
\centering
\vspace{-5pt}
\includegraphics[width=\linewidth]{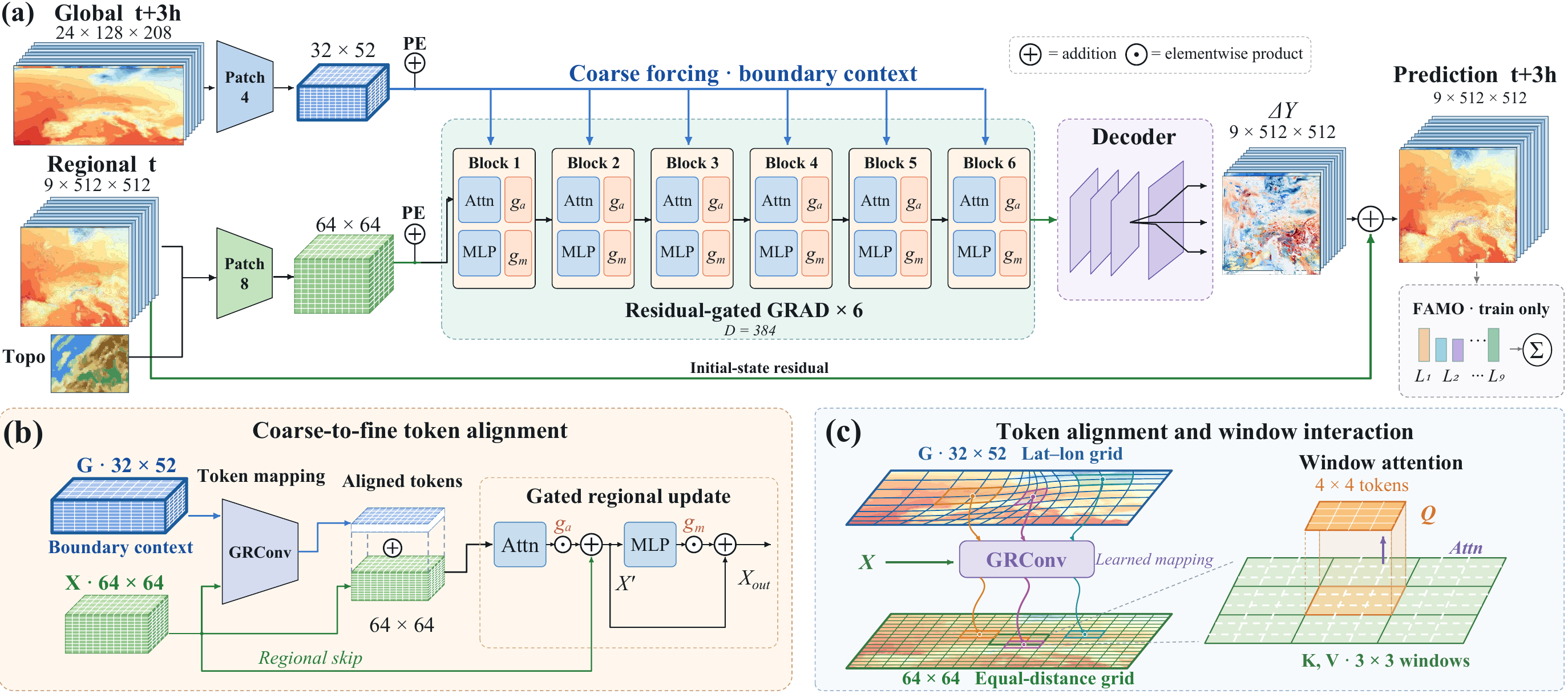}
\caption{\textbf{ScaleCast framework during training.}
        (a) Training pipeline from global analysis at $t+3$~h, the regional state at $t$, and terrain to the regional prediction at $t+3$~h through feature encoding, stacked GRAD blocks, and increment decoding.
        (b) A Global--Regional Alignment and Dynamics (GRAD) block from (a), combining GRConv alignment, regional attention, and gated updates.
        (c) Details of cross-grid token alignment and windowed attention within (b).}
\vspace{-8pt}
\label{fig:architecture}
\end{figure}

\vspace{-5pt}
\paragraph{Problem definition}
\label{sec:problem}
\label{sec:recursive_procedure}
Limited-area forecasts require evolving large-scale conditions as well as a regional initial state~\citep{warner1997lateral}. Hence, we study kilometer-scale regional weather forecasting under global forecast guidance. The regional atmospheric state is represented by $X_{0}\in\mathbb{R}^{C_r\times H_r\times W_r}$, with static terrain features $S\in\mathbb{R}^{H_r\times W_r}$. The global initial state is denoted by $G_0\in\mathbb{R}^{C_g\times H_g\times W_g}$. Here, $C_r$ and $C_g$ denote the numbers of regional and global variables respectively; $(H_r,W_r)$ and $(H_g,W_g)$ specify the regional and global grid dimensions. An existing global forecasting model $\mathcal M$ with frozen parameters $\Theta$ provides the global forecast trajectory. Our goal is to learn a regional forecasting model $\mathcal F_\theta$:
\begin{gather}
\widehat G 
=
\left(
\widehat G_{3},
\widehat G_{6},
\ldots,
\widehat G_{T}
\right)
=
\mathcal M_\Theta(G_{0}),
\label{eq:global-forecast}
\\
\widehat X
=
\left(
\widehat X_{3},
\widehat X_{6},
\ldots,
\widehat X_{T}
\right)
=
\mathcal F_\theta\!\left(
X_{0},S,\widehat G 
\right),
\label{eq:rollout}
\end{gather}
where $\theta$ denotes the learnable parameters, $\widehat X$ and $\widehat G$ denote the sets of regional and global predictions, respectively. Regional predictions are generated autoregressively, using the global forecast at each step. In this work, we consider forecasts at 3-hour intervals up to 72 hours ahead ($T=72$).

\vspace{-5pt}
\paragraph{Training.}
As shown in Figure~\ref{fig:architecture}(a), training uses the global state $G_{t+3}$, together with the regional state $X_t$ and terrain $S$, to predict the regional target $X_{t+3}$. Unlike inference, the global condition is an analysis field rather than a model-generated forecast. We use single-step supervision with a mean squared error for each regional variable $v$:
\begin{equation}
\ell_v(t)
=
\frac{1}{|\Omega_v|}
\sum_{i\in\Omega_v}
\left(
\widehat{X}_{t+3,v,i}
-
X_{t+3,v,i}
\right)^2, \qquad
\mathcal L=\sum_{v\in\mathcal V_+}q_v\ell_v.
\label{eq:loss}
\end{equation}
where $\Omega_v$ contains the valid regional grid points for variable $v$, and tildes denote standardized values. The set $\mathcal V_+$ includes variables with nonempty $\Omega_v$, and the nonnegative weights $q_v$ sum to one. We adapt FAMO~\citep{liu2023famo} to adjust these weights according to loss progress. Masking and weight-update details are provided in Appendix~\ref{app:famo}.

\vspace{-5pt}
\subsection{Global--regional token alignment}
\label{sec:cross_grid}
\vspace{-5pt}
Global forecasts provide large-scale context, but regional prediction requires this information to be expressed at regional locations~\citep{adamov2025limited}. In our setting, the global latitude--longitude grid and the regional projected grid differ in sampling locations and spacing. Their patch indices therefore do not define a one-to-one correspondence. As shown in Figure~\ref{fig:architecture}(b), we address this mismatch with GRConv, which maps joint global--regional features to regional token positions.

For a forecast step from $t$ to $t+\Delta t$, the inputs are the current regional state $\widehat X_t$, terrain $S$, and the global forecast $\widehat G_{t+\Delta t}$, with $\widehat X_0=X_0$. Separate patch encoders $E_r$ and $E_g$ produce
\begin{equation}
\begin{aligned}
R
&= E_r\!\left([\widehat X_t,S]\right)
   + P_r
   \in\mathbb{R}^{N_r\times d}, \qquad
C= E_g\!\left(\widehat G_{t+\Delta t}\right)
   + P_g
   \in\mathbb{R}^{N_g\times d},
\end{aligned}
\label{eq:embeddings}
\end{equation}
where $[\cdot,\cdot]$ denotes channel-wise concatenation, treating terrain as an additional input channel. The encoders embed non-overlapping patches into a common feature width $d$, with $N_r$ and $N_g$ denoting the regional and global patch counts, respectively. Fixed two-dimensional sinusoidal encodings $P_r$ and $P_g$ identify patch positions within each grid. The global forecast $\widehat G_{t+\Delta t}$ is encoded once, and its tokens $C$ are reused across layers within the same forecast step.

At layer $\ell$, let $R^\ell$ denote the regional tokens and $Z^\ell=\operatorname{LN}(R^\ell)$ their normalized representation. GRConv concatenates $Z^\ell$ and $C$ and projects them onto the regional token positions:
\begin{equation}
\begin{aligned}
B^\ell
&= W_g^\ell[Z^\ell;C]
   + b_g^\ell\mathbf{1}_d^\top, \qquad
U^\ell
= Z^\ell+\operatorname{LN}_g^\ell(B^\ell).
\end{aligned}
\label{eq:grconv}
\end{equation}
Here, $[Z^\ell;C]\in\mathbb{R}^{(N_r+N_g)\times d}$ stacks the two token sequences,
$W_g^\ell\in\mathbb{R}^{N_r\times(N_r+N_g)}$ contains the learned mapping weights, and
$b_g^\ell\in\mathbb{R}^{N_r}$ is broadcast across features by the all-ones vector $\mathbf{1}_d$.
Each row of $W_g^\ell$ maps the joint tokens to one regional position. Adding the normalized projection to $Z^\ell$ preserves a direct regional feature path, yielding $U^\ell\in\mathbb{R}^{N_r\times d}$ in the regional token layout. Forecast supervision learns position-specific token combinations aligned with the regional layout. Residual fusion retains regional features while providing global guidance.

\subsection{Regional interaction and state update}
\label{sec:controlled_evolution}
With global guidance incorporated into $U^\ell$, regional prediction further requires modeling interactions between local conditions and their surroundings. GRAD uses input-dependent windowed attention to combine these aligned features with the surrounding regional context.
We partition $U^\ell$ into windows of $b^2$ tokens and denote the features in query window $w$ by $U_w^\ell\in\mathbb{R}^{b^2\times d}$. The gathered key/value features are denoted by $U_{\mathcal N(w)}^\ell\in\mathbb{R}^{9b^2\times d}$. With $H$ attention heads and $d_h=d/H$, head $h$ computes:
\begin{equation}
\begin{aligned}
Q_{w,h} =  U_w^\ell W_{Q,h}^\ell, \quad
K_{w,h} &= U_{\mathcal N(w)}^\ell W_{K,h}^\ell+E_h^\ell,\quad
V_{w,h} =  U_{\mathcal N(w)}^\ell W_{V,h}^\ell,\\
O_{w,h} &= \operatorname{softmax}_{\mathrm{keys}}\!\left(
\frac{Q_{w,h}K_{w,h}^{\top}}{\sqrt{d_h}}
\right)V_{w,h}.
\end{aligned}
\label{eq:neighborhood_attention}
\end{equation}
Here, $W_{Q,h}^\ell,W_{K,h}^\ell,W_{V,h}^\ell\in\mathbb{R}^{d\times d_h}$ are learned projections, and $E_h^\ell\in\mathbb{R}^{9b^2\times d_h}$ encodes positional information for the keys. Scaled dot-product attention~\citep{vaswani2017attention} normalizes each query over the gathered entries, yielding $O_{w,h}\in\mathbb{R}^{b^2\times d_h}$. Concatenating heads, restoring the regional token order, and applying an output projection produces $A^\ell\in\mathbb{R}^{N_r\times d}$. Attention adaptively weights the surrounding fused features to update each regional query.

To incorporate the attention output $A^\ell$ while retaining the regional features $R^\ell$, GRAD uses input-dependent, channel-wise gates for the attention and MLP residual branches~\citep{laitenberger2026gateskip}. We define $g_j^\ell(R)=\sigma(RW_j^\ell+\mathbf{1}_{N_r}(b_j^\ell)^\top)$ for $j\in\{A,M\}$, corresponding to the attention and MLP branches, respectively. The regional features are updated as:
\begin{equation}
\begin{aligned}
R^{\ell+1/2} &= R^\ell+g_A^\ell(R^\ell)\odot A^\ell,\quad
R^{\ell+1} = R^{\ell+1/2}+g_M^\ell(R^{\ell+1/2})\odot M_\ell\!\left(\operatorname{LN}(R^{\ell+1/2})\right).
\end{aligned}
\label{eq:residual_gates}
\end{equation}
Here, $W_j^\ell\in\mathbb{R}^{d\times d}$ and $b_j^\ell\in\mathbb{R}^{d}$ are learned gate parameters, $\sigma$ is sigmoid, $\odot$ denotes element-wise multiplication, and $M_\ell$ is a two-layer GELU MLP. Each gate controls the residual contribution at each position and feature channel while preserving the identity path. The MLP gate uses the attention-updated features $R^{\ell+1/2}$, so its modulation reflects the preceding regional interaction.

\vspace{-5pt}
\paragraph{Decoder.}
Predicting changes makes persistence an explicit reference, following residual parameterization~\citep{he2016resnet}. A shared upsampler and variable-specific heads decode $D_\theta$ to the regional grid. With training statistics $\mu_r,\sigma_r$, the standardized and physical predictions are:
\begin{equation}
\widehat{ X}_{t+\Delta t}=\widehat{X}_t+D_\theta(R^L),\qquad
\widehat X_{t+\Delta t}=\mu_r+\sigma_r\odot\widehat{X}_{t+\Delta t}.
\label{eq:increment}
\end{equation}
These variable-specific heads decode the shared representation using a common spatial upsampler. The decoder includes small variable-specific low-rank branches, detailed in Appendix~\ref{app:decoder}.


\section{Experiments}
\label{sec:experiments}
\vspace{-5pt}
\subsection{Experimental setup}
\vspace{-5pt}
\paragraph{Datasets}
\label{sec:exp_data}
We construct a regional weather dataset integrating global atmospheric analyses, high-resolution regional reanalyses, global forecasts, and station observations. \textbf{ERA5} atmospheric states~\citep{hersbach2020era5} are paired by valid time with \textbf{CERRA} regional targets~\citep{ridal2024cerra} at 3-hour intervals. The regional domain covers a European crop on a $512\times512$ grid at approximately 5.5~km spacing, with static topography provided on the same grid. We use 24 coarse atmospheric variables to predict nine regional variables, as summarized in
Table~\ref{tab:data}. The available reanalysis archive spans 1985--2023. The current experiments use 2023 samples for one-step validation and all samples in 2022 for test. Finally, \textbf{HadISD} temperature and relative-humidity observations and the \textbf{Hodges} cyclone catalogue provide additional references for station evaluation. We also test cross-dataset generalization by transferring the model to \textbf{HRRR}.

\vspace{-5pt}
\paragraph{Metrics}
We use Root Mean Square Error (\textbf{RMSE}), Mean Absolute Error (\textbf{MAE}), and signed \textbf{Bias} to assess gridded forecast accuracy against CERRA reanalysis, and Normalized Mean Square Error (\textbf{nMSE}) to compare multi-variable performance across model variants. Cyclone position error, central-pressure error, and track coverage assess storm prediction against the Hodges catalogue. Station RMSE and MAE evaluate local temperature and relative-humidity forecasts against HadISD observations. Metric definitions and verification procedures are provided in Appendix~\ref{app:eval}.

\vspace{-5pt}
\paragraph{Baselines}
We compare ScaleCast with the numerical forecasts from \textbf{ECMWF-HRES} and \textbf{CERRA NWP}, as well as the global data-driven models \textbf{Pangu-Weather}~\citep{bi2023pangu} and \textbf{GraphCast}~\citep{lam2023graphcast}. Global forecasts are spatially mapped to the regional grid for direct comparison. We further include \textbf{ML-LAM}~\citep{oskarsson2023graph}, \textbf{Hi-LAM}~\citep{oskarsson2024probabilistic}, \textbf{Graph-EFM}~\citep{oskarsson2024probabilistic}, \textbf{OneForecast}~\citep{gao2025oneforecast}, and \textbf{StormCast}~\citep{pathak2026stormcast} as regional forecasting baselines. Graph-EFM uses a four-member ensemble mean, while the other archived regional configurations use one member. We also evaluate ScaleCast with Pangu-Weather, GraphCast, and HRES as alternative global drivers, keeping the regional model weights and normalization fixed. 

\vspace{-5pt}
\paragraph{Implementation}
Our framework is implemented in PyTorch and trained for 100K steps on four NVIDIA A800 GPUs. We use the AdamW optimizer with a peak learning rate of $1\times10^{-4}$, $(\beta_1,\beta_2)=(0.9,0.999)$, and a weight decay of $0.05$. The learning-rate schedule uses three warm-up epochs followed by cosine decay, with a minimum learning rate of $1\times10^{-6}$. The per-device batch size is two, with four gradient-accumulation steps, giving an effective batch size of 32. The model contains six GRAD blocks with a hidden dimension of 384 and eight attention heads. Sparse FAMO balances the variable-specific training objectives. During inference, the model predicts regional states at 3-hour intervals and feeds each prediction into the next step. The same checkpoint is applied to forecasts from \textbf{Pangu-Weather}, \textbf{GraphCast}, and \textbf{ECMWF-HRES} without driver-specific retraining. Additional implementation details are provided in Appendix~\ref{app:impl}.

\definecolor{rollbestcell}{HTML}{DDF2EA}
\newcommand{\rollbest}[1]{\cellcolor{rollbestcell}\textbf{#1}}
\newcommand{\rollsecond}[1]{\underline{#1}}
\newcommand{\rollmetric}[2]{#1}

\begin{table*}[t]
  \centering
  \setlength{\belowcaptionskip}{6pt}
  \vspace{-14pt}
  \caption{\textbf{Recursive regional forecast errors through $+30$~h.} Pixel-pooled RMSE over ten 3-hour leads in 2022. Lower is better.}
  \label{tab:v25-v33-cerra-30h}
  \begin{minipage}{\textwidth}
    \setlength{\tabcolsep}{1pt}
    \renewcommand{\arraystretch}{1.06}
    \footnotesize
    \begin{tabular*}{\textwidth}{@{\extracolsep{\fill}}lccccccccc@{}}
      \toprule
      & \multicolumn{3}{c}{Surface RMSE $\downarrow$} & \multicolumn{6}{c}{Upper-air RMSE $\downarrow$} \\
      \cmidrule(lr){2-4}\cmidrule(l){5-10}
      Method
      & \shortstack{MSLP }
      & \shortstack{T2m }
      & \shortstack{RH2m }
      & \shortstack{T500 }
      & \shortstack{T850}
      & \shortstack{Z500}
      & \shortstack{Z850}
      & \shortstack{RH500}
      & \shortstack{RH850} \\
      \midrule
      ECMWF-HRES  & \rollmetric{92.9}{-4.3\%} & \rollmetric{1.331}{+36.5\%} & \multicolumn{1}{c}{--} & \rollmetric{0.733}{-7.6\%} & \rollmetric{0.828}{+11.2\%} & \rollmetric{0.772}{-3.0\%} & \rollmetric{0.711}{-5.2\%} & \rollmetric{22.680}{+28.9\%} & \rollmetric{12.622}{+18.3\%} \\
      CERRA NWP & \rollmetric{97.1}{\textit{ref}} & \rollmetric{0.975}{\textit{ref}} & \rollmetric{6.732}{\textit{ref}} & \rollmetric{0.793}{\textit{ref}} & \rollsecond{\rollmetric{0.744}{\textit{ref}}} & \rollmetric{0.796}{\textit{ref}} & \rollmetric{0.750}{\textit{ref}} & \rollmetric{17.596}{\textit{ref}} & \rollmetric{10.668}{\textit{ref}} \\
      \addlinespace[2pt]
      Pangu  & \rollmetric{93.9}{-3.3\%} & \rollmetric{1.461}{+49.8\%} & \multicolumn{1}{c}{--} & \rollmetric{0.725}{-8.6\%} & \rollmetric{0.840}{+12.9\%} & \rollmetric{0.775}{-2.6\%} & \rollmetric{0.685}{-8.7\%} & \multicolumn{1}{c}{--} & \multicolumn{1}{c}{--} \\
      GraphCast  & \rollmetric{91.8}{-5.4\%} & \rollmetric{1.431}{+46.7\%} & \multicolumn{1}{c}{--} & \rollsecond{\rollmetric{0.695}{-12.4\%}} & \rollmetric{0.784}{+5.3\%} & \rollmetric{0.733}{-7.9\%} & \rollmetric{0.671}{-10.6\%} & \multicolumn{1}{c}{--} & \multicolumn{1}{c}{--} \\
      \addlinespace[2pt]
      ML-LAM & \rollmetric{95.1}{-2.0\%} & \rollmetric{1.133}{+16.2\%} & \rollmetric{7.347}{+9.1\%} & \rollmetric{0.725}{-8.6\%} & \rollmetric{0.844}{+13.4\%} & \rollmetric{0.851}{+6.9\%} & \rollmetric{0.694}{-7.5\%} & \rollmetric{17.416}{-1.0\%} & \rollmetric{10.908}{+2.2\%} \\
      Hi-LAM & \rollmetric{90.0}{-7.3\%} & \rollmetric{1.041}{+6.8\%} & \rollmetric{7.072}{+5.0\%} & \rollmetric{0.721}{-9.0\%} & \rollmetric{0.795}{+6.9\%} & \rollmetric{0.762}{-4.2\%} & \rollmetric{0.684}{-8.9\%} & \rollmetric{16.953}{-3.7\%} & \rollmetric{10.408}{-2.4\%} \\
      Graph-EFM & \rollmetric{93.7}{-3.4\%} & \rollmetric{1.215}{+24.6\%} & \rollmetric{7.676}{+14.0\%} & \rollmetric{0.729}{-8.0\%} & \rollmetric{0.871}{+17.0\%} & \rollmetric{0.759}{-4.6\%} & \rollmetric{0.738}{-1.6\%} & \rollmetric{17.171}{-2.4\%} & \rollmetric{11.049}{+3.6\%} \\
      OneForecast-Reg & \rollmetric{91.3}{} & \rollmetric{1.408}{} & \multicolumn{1}{c}{--} & \rollmetric{0.735}{} & \rollmetric{0.762}{} & \rollmetric{0.754}{} & \rollmetric{0.692}{} & \multicolumn{1}{c}{--} & \multicolumn{1}{c}{--} \\
      StormCast-Reg & \rollmetric{90.5}{-6.8\%} & \rollmetric{1.004}{+3.0\%} & \rollmetric{6.691}{-0.6\%} & \rollmetric{0.713}{-10.1\%} & \rollmetric{0.760}{+2.2\%} & \rollmetric{0.735}{-7.6\%} & \rollmetric{0.668}{-10.9\%} & \rollsecond{\rollmetric{16.622}{-5.5\%}} & \rollsecond{\rollmetric{9.905}{-7.1\%}} \\
      \addlinespace[2pt]
      \hline
      SC + Pangu & \rollmetric{89.9}{-7.4\%} & \rollmetric{0.958}{-1.8\%} & \rollmetric{6.734}{<+0.1\%} & \rollmetric{0.700}{-11.7\%} & \rollmetric{0.748}{+0.5\%} & \rollmetric{0.735}{-7.7\%} & \rollbest{\rollmetric{0.653}{-12.9\%}} & \rollmetric{16.629}{-5.5\%} & \rollmetric{9.908}{-7.1\%} \\
      SC + GraphCast & \rollbest{\rollmetric{88.2}{-9.1\%}} & \rollbest{\rollmetric{0.926}{-5.1\%}} & \rollsecond{\rollmetric{6.689}{-0.6\%}} & \rollbest{\rollmetric{0.683}{-13.8\%}} & \rollbest{\rollmetric{0.732}{-1.7\%}} & \rollbest{\rollmetric{0.724}{-9.0\%}} & \rollsecond{\rollmetric{0.654}{-12.9\%}} & \rollbest{\rollmetric{16.365}{-7.0\%}} & \rollbest{\rollmetric{9.755}{-8.6\%}} \\
      SC + HRES & \rollsecond{\rollmetric{89.8}{-7.5\%}} & \rollsecond{\rollmetric{0.942}{-3.4\%}} & \rollbest{\rollmetric{6.626}{-1.6\%}} & \rollmetric{0.698}{-11.9\%} & \rollmetric{0.745}{<+0.1\%} & \rollsecond{\rollmetric{0.729}{-8.4\%}} & \rollmetric{0.673}{-10.3\%} & \rollmetric{16.792}{-4.6\%} & \rollmetric{10.145}{-4.9\%} \\
      \bottomrule
    \end{tabular*}
  \end{minipage}
  \vspace{-15pt}
\end{table*}


\begin{figure}[t]
\centering
\vspace{-8pt}
\includegraphics[width=\linewidth]{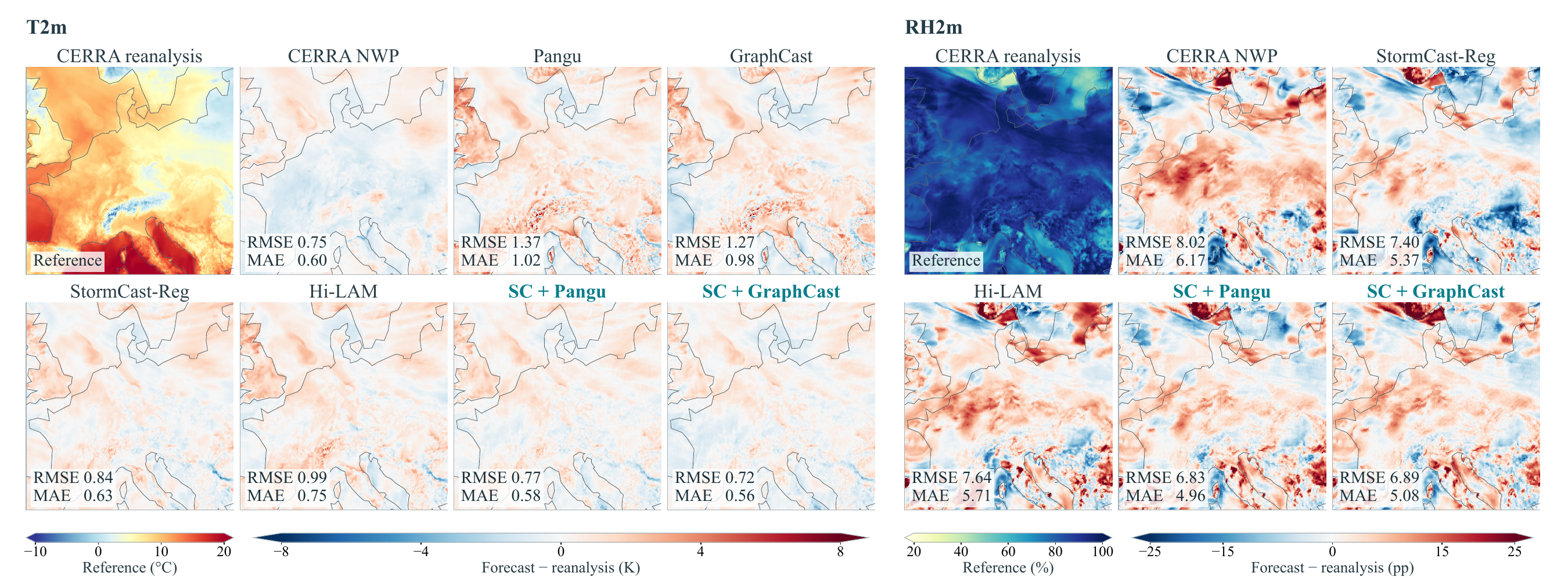}
\vspace{-16pt}
\caption{Regional errors for two cases: T2m at $+9\,$h from 16 November 2022, 00 UTC (left), and RH2m at $+27\,$h from 5 April 2022, 00 UTC (right). CERRA shows absolute values; forecasts show deviations from CERRA. Pangu and GraphCast RH2m are unavailable and omitted.}
\vspace{-6pt}
\label{fig:regional-error}
\end{figure}

\vspace{-8pt}
\subsection{Overall Performance}
\label{sec:exp_baselines}
\paragraph{Comparison with forecasting baselines}
We compare ScaleCast with numerical, global data-driven, and regional forecasting models on nine variables in Table~\ref{tab:v25-v33-cerra-30h}. Results report pixel-pooled RMSE across ten leads through $+30\,$h. More Results in Appendix~\ref{app:driver-robustness}. We observe the following:

\begin{itemize}[leftmargin=*]
\vspace{-4pt}
\item With GraphCast, ScaleCast outperforms all non-ScaleCast baselines on nine variables. Compared with StormCast-Reg, it reduces T2m, T500, and T850 RMSE by approximately $7.8\%$, $4.2\%$, and $3.7\%$, respectively, improving temperature prediction near the surface and aloft.

\item ScaleCast improves each global driver's forecasts across all available variables, reducing T2m RMSE by $0.503$, $0.505$, and $0.389$~K with Pangu-Weather, GraphCast, and HRES, respectively. These gains demonstrate added regional forecast skill beyond regridding global forecasts.

\item With GraphCast, ScaleCast outperforms CERRA NWP on all nine variables, reducing MSLP and T500 RMSE by $8.9$~Pa and $0.110$~K, respectively. RH500 and RH850 RMSE also decrease by approximately $7.0\%$ and $8.6\%$, showing gains across pressure, temperature, and humidity.

\item Among the ScaleCast configurations, GraphCast guidance gives the lowest errors on seven variables, while HRES leads on RH2m and Pangu-Weather on Z850. With one shared checkpoint and no driver-specific retraining, ScaleCast improves on all three global drivers.

\end{itemize}
\vspace{-5pt}
\paragraph{Regional visualization}
Figure~\ref{fig:regional-error} compares T2m and RH2m errors in two selected cases. ScaleCast reduces the widespread T2m warm biases in Pangu-Weather and GraphCast. With GraphCast, it achieves the lowest T2m RMSE of $0.72$~K, improving on GraphCast and CERRA NWP by $0.55$ and $0.03$~K, respectively. For RH2m, ScaleCast reduces broad positive and localized negative biases; the Pangu-driven model achieves an RMSE of $6.83$ percentage points, below CERRA NWP and StormCast. These case-specific gains indicate closer agreement with CERRA, although localized errors remain. Absolute fields are provided in Appendix~\ref{app:regional-fields}.

\vspace{-5pt}
\subsection{Module comparisons and ablations}


\definecolor{ablationbest}{HTML}{DDF2EA}
\providecommand{\abest}[1]{\cellcolor{ablationbest}\textbf{#1}}
\providecommand{\asecond}[1]{\underline{#1}}

\begin{table*}[t]
  \centering
  \vspace{-14pt}
  \caption{\textbf{Module ablations.}
  Bold and underlined values indicate the best and second-best
  results.}
  \label{tab:module-replacement-ablation}
  \vspace{6pt}
  \begin{threeparttable}
    \setlength{\tabcolsep}{1.45pt}
    \renewcommand{\arraystretch}{1.23}
    \scriptsize
    \resizebox{\textwidth}{!}{%
    \begin{tabular}{@{}llrrrrrrrrrr@{}}
      \toprule
      \multicolumn{2}{c}{Design choices}
      & \multicolumn{3}{c}{Surface RMSE}
      & \multicolumn{6}{c}{Upper-air RMSE}
      & \\
      \cmidrule(lr){1-2}
      \cmidrule(lr){3-5}
      \cmidrule(lr){6-11}
      Type & Module
      & \shortstack{MSLP $\downarrow$\\(Pa)}
      & \shortstack{T2m $\downarrow$\\(K)}
      & \shortstack{RH2m $\downarrow$\\(pp)}
      & \shortstack{T500 $\downarrow$\\(K)}
      & \shortstack{T850 $\downarrow$\\(K)}
      & \shortstack{Z500 $\downarrow$\\(dam)}
      & \shortstack{Z850 $\downarrow$\\(dam)}
      & \shortstack{RH500 $\downarrow$\\(pp)}
      & \shortstack{RH850 $\downarrow$\\(pp)}
      & \shortstack{nMSE $\downarrow$\\($10^{-2}$)} \\
      \midrule

      \multirow{4}{*}{\textit{Attention}}
      & Differential
      & 41.1 & 0.724 & \abest{5.203}
      & 0.379 & 0.459 & 0.477 & 0.341
      & 7.489 & 6.105 & 2.733 \\

      & Gated
      & 36.7 & \abest{0.705} & 5.226
      & \asecond{0.338} & \abest{0.427} & 0.429 & 0.309
      & \asecond{7.420} & \asecond{5.931} & \asecond{2.713} \\

      & Cross-attention
      & \asecond{35.9} & \asecond{0.708} & \asecond{5.219}
      & \abest{0.337} & 0.430 & \asecond{0.410} & \asecond{0.298}
      & 7.623 & 6.020 & 2.765 \\

      & Our attention
      & \abest{34.6} & 0.710 & 5.233
      & 0.346 & \asecond{0.428} & \abest{0.409} & \abest{0.291}
      & \abest{7.396} & \abest{5.893} & \abest{2.702} \\

      \midrule

      \multirow{4}{*}{\textit{Spatial}}
      & Scale + terrain
      & 62.7 & 0.931 & 5.484
      & 0.510 & 0.541 & 0.657 & 0.512
      & 9.337 & 6.652 & 3.579 \\

      & Geo-nested
      & 41.6 & \asecond{0.761} & \asecond{5.318}
      & 0.392 & 0.465 & 0.486 & 0.350
      & 8.242 & 6.279 & 3.033 \\

      & Geo-remap + fusion
      & \asecond{40.8} & 0.766 & 5.336
      & \asecond{0.383} & \asecond{0.462}
      & \asecond{0.476} & \asecond{0.336}
      & \asecond{8.173} & \asecond{6.217} & \asecond{3.009} \\

      & GRConv
      & \abest{37.9} & \abest{0.742} & \abest{5.294}
      & \abest{0.361} & \abest{0.447}
      & \abest{0.461} & \abest{0.302}
      & \abest{7.835} & \abest{6.023} & \abest{2.872} \\

      \midrule

      \multirow{4}{*}{\textit{Decoder}}
      & Direct prediction
      & 51.7 & 0.835 & 5.443
      & 0.463 & 0.554 & 0.652 & 0.408
      & 8.959 & 7.190 & 3.546 \\

      & Separate
      & 41.6 & 0.761 & 5.318
      & 0.392 & 0.465 & 0.486 & 0.350
      & \asecond{8.242} & 6.279 & 3.033 \\

      & Shared
      & \abest{38.0} & \asecond{0.724} & \abest{5.260}
      & \asecond{0.375} & \asecond{0.451}
      & \asecond{0.454} & \abest{0.320}
      & 8.317 & \asecond{6.264} & \asecond{2.996} \\

      & Low-rank + res.
      & \asecond{39.3} & \abest{0.723} & \asecond{5.283}
      & \abest{0.374} & \abest{0.450}
      & \abest{0.450} & \asecond{0.328}
      & \abest{8.196} & \abest{6.228} & \abest{2.974} \\

      \bottomrule
    \end{tabular}%
    }
  \end{threeparttable}
\end{table*}

\vspace{-4pt}
Table~\ref{tab:module-replacement-ablation} compares spatial alignment, attention variants, and decoder designs under the single-step validation setting. We observe the following:

\begin{itemize}[leftmargin=*]
\vspace{-4pt}
\item Our attention achieves  the lowest errors in MSLP, geopotential height, and upper-air humidity. Gated attention performs best on T2m and T850, while cross-attention leads on T500. The proposed design therefore provides variable-specific advantages for alternative attention mechanisms.

\item GRConv outperforms scale-and-terrain conditioning, Geo-nested coupling, and geographic remapping with fusion on all nine variables. These gains favor learned token alignment over geographic remapping and fusion.

\item Shared decoding reduces T2m RMSE by $0.037$~K relative to separate heads. Low-rank residual heads further improve six of nine variables over shared decoding. The resulting decoder also outperforms direct prediction on all nine variables, supporting the combined use of shared features, lightweight variable-specific heads, and residual prediction.
\end{itemize}

\vspace{-4pt}
Additional comparisons of spatial training strategies and humidity objectives are provided in Appendix~\ref{app:training-comparisons}. The effect of expanding coarse-input coverage is examined in Appendix~\ref{app:input-coverage}.

\vspace{-4pt}
\subsection{Case Study }
\begin{figure*}[t]
  \centering
  \vspace{-5pt}
  \includegraphics[width=\textwidth]{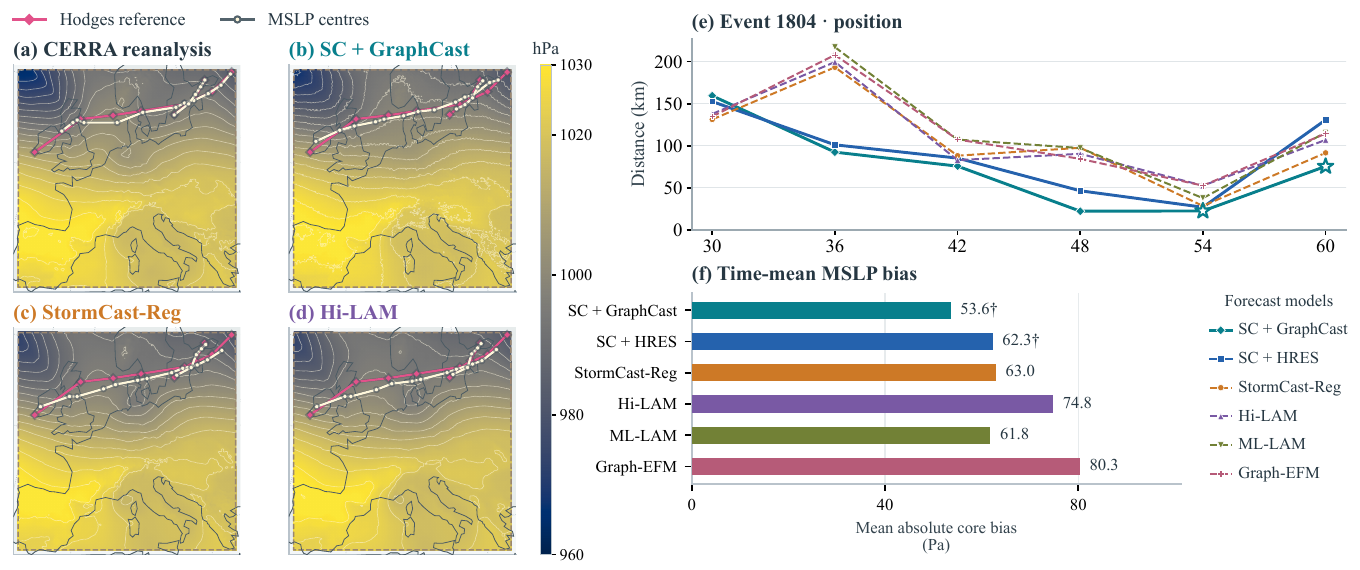}
  \vspace{-12pt}
  \caption{
  \textbf{Windstorm position and core-pressure diagnostics.}
  (a--d) MSLP at $+72$~h and cyclone tracks. (e) Position errors for event 1804. (f) Time-mean absolute core-pressure bias relative to CERRA within $200$~km of Hodges centres.
  }
  \label{fig:windstorm-selected}
  \vspace{-8pt}
\end{figure*}

\vspace{-4pt}
\paragraph{Extratropical cyclone tracking}
\label{sec:exp_windstorm}
Figure~\ref{fig:windstorm-selected} examines whether regional forecast gains extend to cyclone position and core pressure in a windstorm case. Although the models reproduce similar large-scale pressure patterns, their cyclone tracks differ. For event 1804, GraphCast-driven ScaleCast yields the lowest position errors among the displayed forecasts from $+36$ to $+60$~h. The comparison shows that similar pressure patterns can accompany substantial differences in storm location, which ScaleCast captures more accurately in this event.

To assess pressure errors alongside track displacement, we compare the time-mean absolute core bias within $200$~km of the Hodges centres over the same six verification times. GraphCast-driven ScaleCast achieves the lowest reported bias of $53.6$~Pa, improving on StormCast-Reg and Hi-LAM by $9.4$ and $21.2$~Pa, respectively. Together, the track and pressure results demonstrate more accurate storm positioning and smaller core-pressure offsets relative to CERRA in this event, extending the evaluation beyond grid-averaged errors. Results for more events are provided in Appendix~\ref{app:windstorm-candidates}.

\vspace{-4pt}
\paragraph{Temperature and humidity at observed stations}
\label{sec:exp_stations}

\begin{figure*}[tp]
  \centering
  \vspace{-8pt}
  \includegraphics[width=\textwidth]{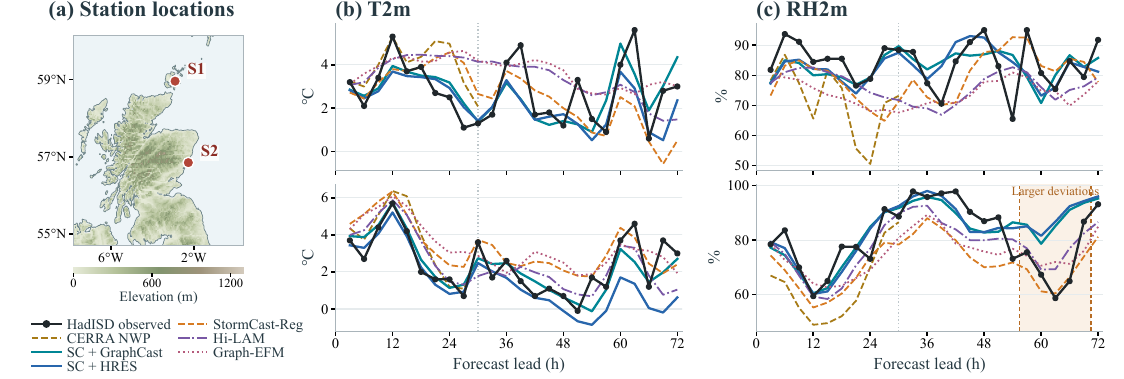}
  \vspace{-20pt}
  \caption{
      \textbf{T2m and RH2m against HadISD stations.}
      (a) Station locations and terrain. (b,c) T2m and RH2m forecasts against HadISD observations through $+72$~h at S1(top) and S2(bottom).
  }
  \label{fig:station-t2m-rh2m}
\end{figure*}


Figure~\ref{fig:station-t2m-rh2m} assesses whether improvements against gridded reanalysis translate into closer agreement with observed local weather. ScaleCast captures the broad cooling after the early temperature peak at both stations. At S2, it follows the observed temperature decline more closely than StormCast-Reg and Graph-EFM during approximately $+18$ to $+30$~h. 
For RH2m, ScaleCast avoids the pronounced CERRA NWP dry bias near $+24$~h at S1 and better captures the high-humidity period from $+30$ to $+42$~h at S2. Agreement with these observed changes demonstrates benefits in local weather evolution. However, both configurations remain too humid during the sharp drying at S2 around $+57$--$+69$~h, indicating that capturing abrupt local transitions remains challenging. Spatial station-error maps are provided in Appendices~\ref{app:station-spatial}.

\begin{figure*}[tp]
  \centering
  \vspace{-8pt}
  \includegraphics[width=\textwidth]{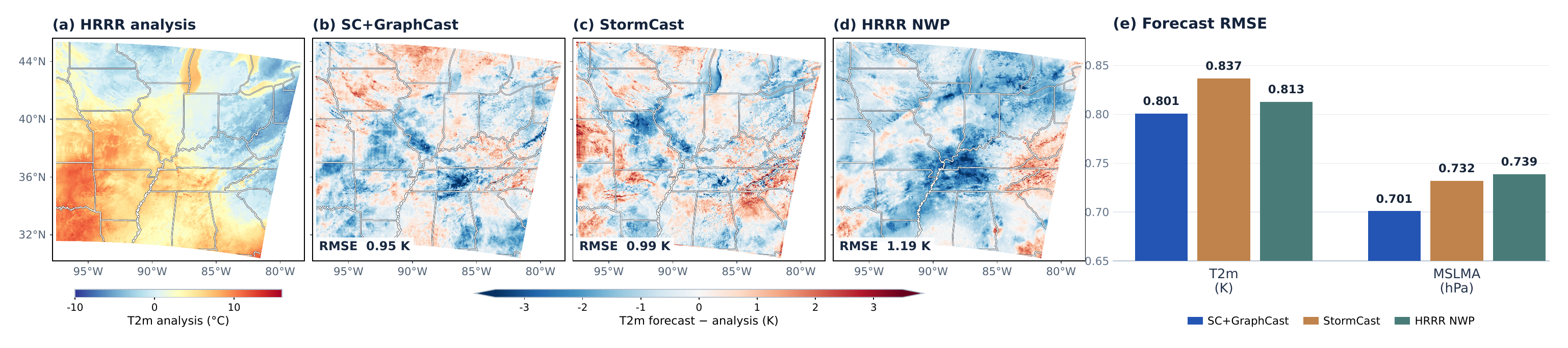}
  \vspace{-20pt}
  \caption{\textbf{Cross-dataset experiment results.}
      (a) HRRR analysis of 2 m temperature. (b--d) Models' forecasts minus the same analysis. (e) RMSE comparison for T2m and MSLMA.
  }
  \label{fig:data-migration}
  \vspace{-8pt}
\end{figure*}

\vspace{-4pt}
\paragraph{Cross-dataset evaluation on HRRR.} 
We assess cross-dataset transfer by fine-tuning the CERRA-trained ScaleCast model on an HRRR subdomain. At \(+3\) h, SC+GraphCast reduces T2m and MSLMA RMSE by \(4.3\%\) and \(4.2\%\) relative to StormCast, and by \(1.5\%\) and \(5.1\%\) relative to HRRR NWP, respectively. The signed-error maps illustrate the spatial differences among the forecasts. The gain over HRRR NWP is modest for T2m but more pronounced for MSLMA, indicating that transfer performance varies by variable. Because the learned models use ERA5 fields valid at the forecast time, this experiment assesses cross-dataset transfer rather than information-matched operational skill. Additional visualizations in Figure~\ref{fig:hrrr_visual} of the Appendix.


\vspace{-4pt}
\section{Conclusion}
\vspace{-4pt}
We present \model{} for kilometer-scale regional forecasting through \emph{Global--Regional Alignment}. GRConv maps joint global--regional tokens to regional positions, while GRAD combines this alignment with windowed attention and gated residual updates. Together with increment decoding, these modules integrate global guidance into the autoregressive evolution of an existing regional state. Trained on ERA5--CERRA pairs, the regional model is reused with Pangu-Weather, GraphCast, and HRES without driver-specific retraining. Evaluated 30-hour rollouts show improved surface and upper-air forecasts, with the GraphCast-driven configuration outperforming CERRA NWP on all nine variables. Component comparisons reveal useful design choices and variable-dependent trade-offs. Windstorm and station case studies further demonstrate improvements in cyclone positioning, core pressure, and selected local temperature and humidity changes. Experiments on HRRR further demonstrate the framework's cross-dataset transferability. Evaluation across additional regions and forecasting systems would help establish broader applicability. These results support using learned global--regional alignment to improve regional forecasts.

\label{maintext:end}

\section*{AI Use Statement}
Generative AI tools were used to assist with manuscript drafting and polishing, consistency checks, literature discovery, implementation inspection, and the preparation of analysis code, figures, and LaTeX. AI-generated text was not treated as scientific evidence or cited as a source. The authors retain responsibility for verifying the cited literature and for the accuracy and integrity of the manuscript, including its methodology, empirical claims, numerical results, and references.

\section*{Ethics Statement}
\model{} is intended for research on regional weather forecasting. The study evaluates regional forecasts over Europe and examines cross-dataset transfer through fine-tuning on HRRR, with additional verification against station observations and windstorm reference tracks. Forecast errors and dependence on upstream global guidance should be considered before operational use, particularly for severe-weather warnings and other safety-critical decisions. The reported results do not constitute operational certification or replace established warning procedures.

\section*{Reproducibility Statement}
Sections~\ref{sec:method} and~\ref{sec:experiments} describe the forecasting formulation, model components, datasets, baselines, and experimental settings. Appendix~\ref{app:impl} specifies the reference architecture, with implementation and training details provided in the following subsections. Appendix~\ref{app:eval} defines the atmospheric variables, error metrics, and aggregation procedures. Appendix~\ref{app:supplementary-experiments} provides lead-dependent results, global-driver comparisons, station and windstorm diagnostics, and additional design comparisons. Data sources and baseline methods are identified through citations.


\clearpage
\appendix

\raggedbottom
\setlength{\parskip}{3pt plus 1pt minus 1pt}
\setlength{\floatsep}{8pt plus 2pt minus 2pt}
\setlength{\textfloatsep}{10pt plus 2pt minus 2pt}
\setlength{\intextsep}{8pt plus 2pt minus 2pt}
\setlength{\dblfloatsep}{8pt plus 2pt minus 2pt}
\setlength{\dbltextfloatsep}{10pt plus 2pt minus 2pt}
\setlength{\abovecaptionskip}{4pt}
\setlength{\belowcaptionskip}{2pt}
\renewcommand{\topfraction}{0.95}
\renewcommand{\bottomfraction}{0.95}
\renewcommand{\textfraction}{0.05}
\renewcommand{\floatpagefraction}{0.70}
\renewcommand{\dbltopfraction}{0.95}
\renewcommand{\dblfloatpagefraction}{0.70}
\setcounter{topnumber}{5}
\setcounter{bottomnumber}{5}
\setcounter{totalnumber}{8}
\setcounter{dbltopnumber}{5}

\makeatletter
\newcommand{\appendixblockbegin}[1]{%
  \par\addvspace{6pt}%
  \noindent\begin{minipage}{\linewidth}%
  \def\@captype{#1}%
}
\newcommand{\appendixblockend}{%
  \end{minipage}\par\addvspace{6pt}%
}
\renewenvironment{figure}[1][]{\appendixblockbegin{figure}}{\appendixblockend}
\renewenvironment{figure*}[1][]{\appendixblockbegin{figure}}{\appendixblockend}
\renewenvironment{table*}[1][]{\appendixblockbegin{table}}{\appendixblockend}
\makeatother


\section{Implementation and training details}

\subsection{Reference configuration}
\label{app:impl}

The configured global input has 24 channels on a $128\times208$ grid, and the regional state has nine channels on a $512\times512$ grid. Patch sizes of four and eight yield $N_g=1{,}664$ and $N_r=4{,}096$ tokens, respectively. Six blocks use width $d=384$, eight heads ($d_h=48$) and MLP expansion four. Regional windows have $b=4$, giving a $16\times16$ window lattice. Independent strided convolutions embed the inputs. Fixed sine--cosine positional encodings represent patch indices, not geographic coordinates. Terrain is min--max scaled and concatenated only to the regional encoder input.

\subsection{HRRR transfer setup.}
\begin{figure}[!htbp]
  \centering
  \includegraphics[width=\linewidth]{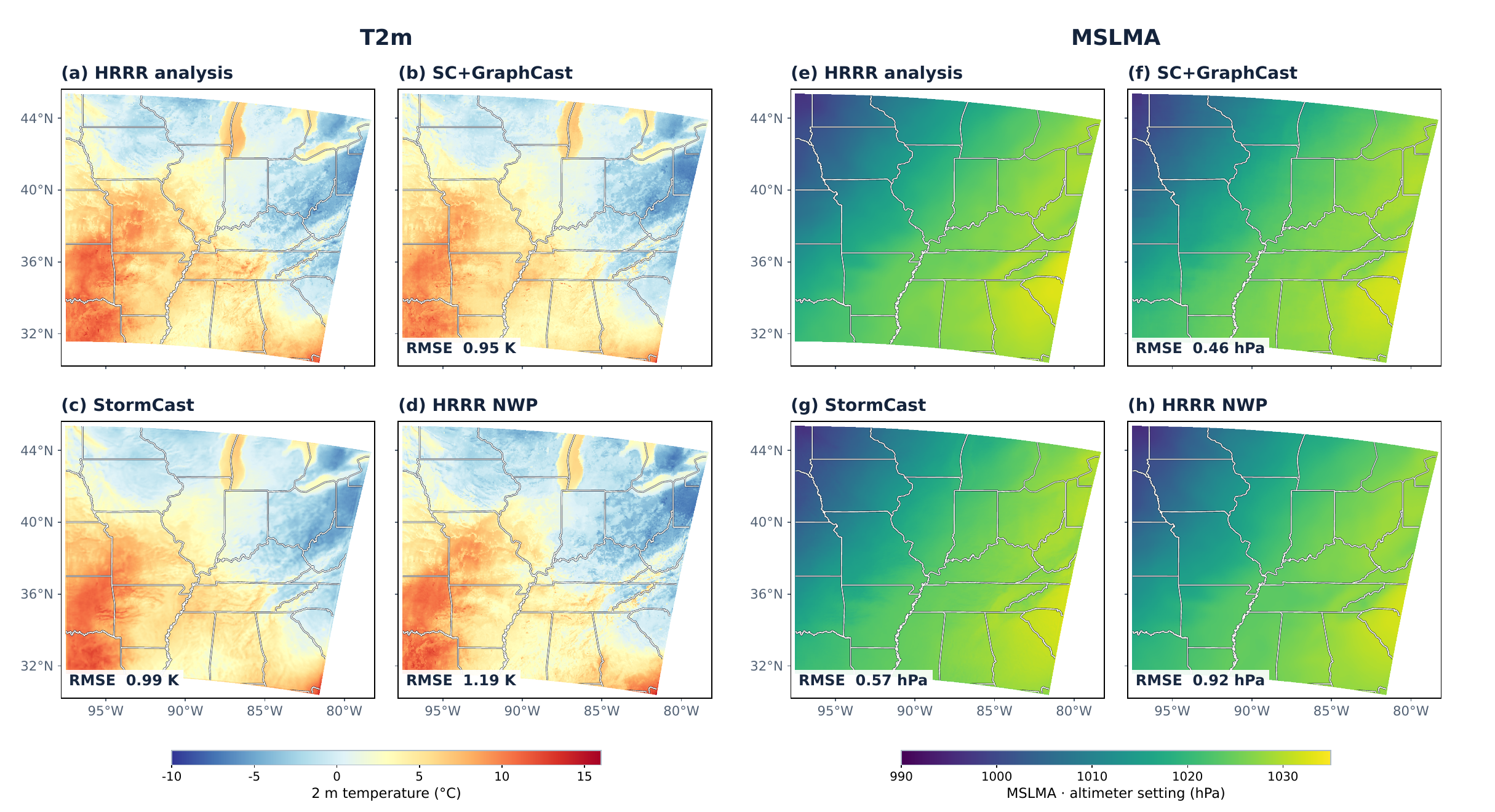}
  \caption{HRRR analysis fields underlying the regional comparisons. Rows show 09 UTC on 2 and 28 December 2022; columns show T2m and MSLMA.}
  \label{fig:hrrr_visual}
\end{figure}

We fine-tuned the validated ERA5-to-CERRA ScaleCast checkpoint on HRRR, replacing the terrain and reinitializing four output heads with different target definitions while retaining the backbone and five compatible heads. Each sample uses HRRR at \(t\) and ERA5 analysis at \(t+3\) h to predict HRRR at \(t+3\) h. The SC+GraphCast label in Figures~\ref{fig:data-migration} and~\ref{fig:hrrr_visual} identifies the transferred regional model; its HRRR conditioning source is GraphCast. Full-model fine-tuning first used January--August 2022, then 2019--2021 and the same 2022 training period with a surface-weighted loss. September--October 2022 served for checkpoint selection and November--December for evaluation.

\subsection{Token mapping, ordered gathering and residual paths}
\label{app:gather}
With tokens in rows, the kernel-size-one GRConv is exactly the affine map in Equation~\ref{eq:grconv}: its convolution channels are token positions, and its sequence coordinate is the feature coordinate. LayerNorm acts on the last, feature dimension. The outer block first normalizes $R^\ell$; attention then adds a separately normalized GRConv result to this normalized regional input. The attention output is subsequently gated and added to the \emph{unnormalized} outer residual $R^\ell$. The MLP gate similarly reads the state after the attention residual, before the MLP's LayerNorm. These are distinct inner and outer residual paths. In this configuration, Q/K normalization, LayerScale and stochastic dropout are disabled. Both gate matrices start at zero with bias two, so the initial gate value is $\operatorname{sigmoid}(2)\simeq0.881$, not a near-zero residual branch.

\paragraph{Regional window interaction.}
Figure~\ref{fig:architecture}(c) illustrates the flow from aligned tokens to windowed attention. The ordered gather $\mathcal N(w)$ in Equation~\ref{eq:neighborhood_attention} specifies the key/value inputs. In the archived implementation, a query window $(i,j)$ lies on an $n_x\times n_y$ lattice with zero-based indices and $n_x,n_y\geq2$. The padding routine prepends both extra slices, giving the index map
\begin{equation}
p_n(s)=
\begin{cases}
n-2,&s=0,\\
0,&s=1,\\
s-2,&2\leq s\leq n+1.
\end{cases}
\label{eq:gather_padding}
\end{equation}
In order, the gathered windows are the query window $(i,j)$ followed by
\begin{equation}
\bigl(p_{n_x}(i+a),p_{n_y}(j+c)\bigr),
\qquad
(a,c)\in\bigl[(0,0),(0,1),(0,2),(1,2),(2,2),(2,1),(2,0),(1,0)\bigr].
\label{eq:gather_indices}
\end{equation}
Each gathered window contributes $b^2$ tokens in row-major order, yielding $9b^2$ key/value entries. The ordered gather can repeat the query and is not uniformly centered. Figure~\ref{fig:architecture}(c) illustrates the interaction mechanism, while the equations here specify the archived indexing.

The learned positional tensor has shape $9\times H\times d_h$. Its rows repeat along the key sequence as $E_h[s,:]=e_h[s\bmod9,:]$ for $s=0,\ldots,9b^2-1$, and are added to keys rather than attention logits or values. The term is thus indexed by key-sequence position, not by neighboring window. Attention outputs are restored to the regional token order.

\subsection{Variable-specific increment decoder}
\label{app:decoder}
Reshape $R^L$ to a patch field $F\in\mathbb{R}^{d\times h_r\times w_r}$, where $(h_r,w_r)$ is the regional patch lattice. For regional patch size $p_r$, the shared full-resolution features and variable increments are
\begin{equation}
\begin{aligned}
H_{\mathrm{dec}} &= \operatorname{GELU}\!\left(\mathcal U(F)\right),\\
[D_\theta(R^L)]_v
&= \operatorname{Conv}_{3\times3}^{v}(H_{\mathrm{dec}})
+\eta\,\operatorname{PixelShuffle}_{p_r}\!\left(\mathcal B_v(F)\right).
\end{aligned}
\label{eq:decoder_detail}
\end{equation}
Here, $\mathcal U$ uses a transposed convolution with kernel size and stride $p_r=8$, reducing the channel width from 384 to 192. Each variable-specific head applies a padded $3\times3$ convolution to produce one output channel. The low-rank branch $\mathcal B_v$ uses two $1\times1$ convolutions with GELU between them, projecting $384\rightarrow8\rightarrow p_r^2$ channels before pixel shuffling, with gain $\eta=0.1$. Its final projection is zero-initialized, leaving the base decoder unchanged at initialization. The decoded increment is added to the standardized regional input (Equation~\ref{eq:increment}).

\subsection{Masked learning and the adapted FAMO controller}
\label{app:famo}
The loss in Equation~\ref{eq:loss} pools squared errors and valid-target counts over the effective batch, including gradient accumulation and distributed workers. 
Let $u$ count successful optimizer updates, $p_v$ be a normalized static prior and $\xi_v$ a controller logit. Denote by $\bar\ell_{v,u-1}$ the lagged moving average available before update $u$, and normalize all expressions over its active set $\mathcal V_+(u)$. After warmup,
\begin{equation}
z_{v,u}=\frac{p_v e^{\xi_v}}{\sum_{j\in\mathcal V_+(u)}p_j e^{\xi_j}},\quad
q_{v,u}=(1-\lambda)
\frac{z_{v,u}/(\bar\ell_{v,u-1}+\epsilon)}
{\sum_{j\in\mathcal V_+(u)}z_{j,u}/(\bar\ell_{j,u-1}+\epsilon)}
+\lambda\frac{p_v}{\sum_{j\in\mathcal V_+(u)}p_j}.
\label{eq:famo_coefficients}
\end{equation}
The implementation evaluates the inverse-loss normalization in log space. Equal priors are configured, but the resulting task coefficients need not be equal. The first 20 successful updates, or an active variable without a previous observation, use the renormalized prior alone. Here $\lambda=0.05$ and $\epsilon=10^{-8}$. After a successful update, the pre-update loss initializes an unseen variable's average; otherwise it updates it as $0.9\bar\ell_{v,u-1}+0.1\ell^{\mathrm{before}}_{v,u}$. Inactive variables retain their previous state.

Every twentieth successful update, beginning at update 20, a no-gradient replay of the same effective batch measures the post-update loss. Following the loss-progress principle of FAMO~\citep{liu2023famo}, define
\begin{equation}
\begin{aligned}
\delta_{v,u}&=\operatorname{clip}\!\left[
\log(\ell^{\mathrm{before}}_{v,u}+\epsilon)-
\log(\ell^{\mathrm{after}}_{v,u}+\epsilon),-1,1\right],\\
g_{v,u}&=z_{v,u}\!\left(\delta_{v,u}-\sum_jz_{j,u}\delta_{j,u}\right)+\gamma\xi_v .
\end{aligned}
\label{eq:famo_probe}
\end{equation}
Inactive entries have $\delta=0$ and are not updated. Controller logits take an Adam step on $g$, with learning rate $0.025$, $\gamma=0.001$, moments $(0.9,0.999)$, numerical epsilon $10^{-8}$ and per-variable step counts. Post-update losses affect the progress signal, not the moving average. Sparse paired probes, lagged loss denominators, clipping, warmup and uniform mixing are this implementation's adaptations; the original FAMO guarantees do not automatically apply. No variable-specific lower weight bounds or additional spectral/gradient losses are configured.


\section{Metrics}
\label{app:eval}

\subsection{Variable inventory and reported field errors}

\begin{table}[!htbp]
    \centering
    \caption{\textbf{Atmospheric variables used by ScaleCast.}
    Global atmospheric inputs contain 24 channels, while the regional
    state and prediction contain nine channels. Pressure levels are
    given in hPa, and physical units are listed before normalization.}
    \label{tab:data}
    \small
    \setlength{\tabcolsep}{4pt}
    \renewcommand{\arraystretch}{1.15}
    \begin{threeparttable}
        \begin{tabular*}{\textwidth}{
            @{\extracolsep{\fill}}lcccc@{}
        }
            \toprule
            Variable
            & Symbol
            & Level / height
            & Unit
            & Channels \\
            \midrule

            \rowcolor{black!5}
            \multicolumn{5}{@{}l}{
                \textbf{Global atmospheric inputs}
                \hfill \textbf{24 channels}
            } \\
            \addlinespace[3pt]

            2-m temperature
            & $T_{2\mathrm{m}}$
            & 2 m
            & K
            & 1 \\

            Mean sea-level pressure
            & $\mathrm{MSLP}$
            & Sea level
            & Pa
            & 1 \\

            10-m zonal wind
            & $u_{10\mathrm{m}}$
            & 10 m
            & $\mathrm{m\,s^{-1}}$
            & 1 \\

            10-m meridional wind
            & $v_{10\mathrm{m}}$
            & 10 m
            & $\mathrm{m\,s^{-1}}$
            & 1 \\

            \addlinespace[3pt]

            Temperature
            & $T$
            & 50, 500, 850, 1000
            & K
            & 4 \\

            Geopotential
            & $z$
            & 50, 500, 850, 1000
            & $\mathrm{m^{2}\,s^{-2}}$
            & 4 \\

            Zonal wind
            & $u$
            & 50, 500, 850, 1000
            & $\mathrm{m\,s^{-1}}$
            & 4 \\

            Meridional wind
            & $v$
            & 50, 500, 850, 1000
            & $\mathrm{m\,s^{-1}}$
            & 4 \\

            Specific humidity
            & $q$
            & 50, 500, 850, 1000
            & $\mathrm{kg\,kg^{-1}}$
            & 4 \\

            \midrule

            \rowcolor{black!5}
            \multicolumn{5}{@{}l}{
                \textbf{Regional state and prediction}
                \hfill \textbf{9 channels}
            } \\
            \addlinespace[3pt]

            Mean sea-level pressure
            & $\mathrm{MSLP}$
            & Sea level
            & Pa
            & 1 \\

            2-m temperature
            & $T_{2\mathrm{m}}$
            & 2 m
            & K
            & 1 \\

            2-m relative humidity
            & $\mathrm{RH}_{2\mathrm{m}}$
            & 2 m
            & \%
            & 1 \\

            \addlinespace[3pt]

            Temperature
            & $T$
            & 500, 850
            & K
            & 2 \\

            Geopotential
            & $z$
            & 500, 850
            & $\mathrm{m^{2}\,s^{-2}}$
            & 2 \\

            Relative humidity
            & $\mathrm{RH}$
            & 500, 850
            & \%
            & 2 \\

            \bottomrule
        \end{tabular*}

    \end{threeparttable}
\end{table}


For a common validity mask $m_{nivk}$ and physical forecast error $e_{nivk}$ for case $n$, grid cell $i$, variable $v$, and lead $k$, the reported per-lead pixel-pooled RMSE is
\begin{equation}
\operatorname{RMSE}_{v,k}=\sqrt{\frac{\sum_{n,i}m_{nivk}e_{nivk}^2}{\sum_{n,i}m_{nivk}}}.
\end{equation}
MAE substitutes $|e|$ and signed bias substitutes $e$ in the same valid-pixel mean. Pooled scores across leads combine their squared errors and valid counts before taking the square root; they are not the unweighted average of ten lead-wise RMSE values. Valid grid cells receive equal weight in these gridded scores. MSLP RMSE is reported in Pa in both the forecast and module-comparison tables; pressure-field maps use hPa as indicated by their colour bars. Relative-humidity differences are in percentage points. Geopotential $z$ and geopotential height $z/g_0$ retain distinct labels, with $g_0=9.80665~\mathrm{m\,s^{-2}}$. No conversion from specific to relative humidity is made without pressure, temperature, and a declared saturation-vapor-pressure convention.

For the one-step module tables, the reported nMSE is the equal-variable mean of per-variable masked MSE in the training-standardized space:
\begin{equation}
\operatorname{nMSE}=\frac{1}{|\mathcal V|}\sum_{v\in\mathcal V}\frac{\sum_{n,i}m_{niv}\bigl[(\widehat Y_{niv}-Y_{niv})/\sigma_v\bigr]^2}{\sum_{n,i}m_{niv}},
\end{equation}
where $\mathcal V$ contains the nine regional variables and $\sigma_v$ is the stored training-set standard deviation. Tables display this dimensionless score in units of $10^{-2}$. The main-table replacement settings are specified in Appendix~\ref{app:module-settings}. The supplementary training and coverage comparisons additionally vary objectives, checkpoint selection, or normalization, as described with their results.


\FloatBarrier
\section{Supplementary experiments and case analyses}
\label{app:supplementary-experiments}
\subsection{Forecast trajectories and global-driver sensitivity}
\label{app:driver-robustness}

The completed comparisons in this section use ten recursive forecast leads. They test how the frozen regional predictor responds to the global driver and how errors evolve with lead time. The absolute-field case at the end complements the error maps in the main text.

\subsubsection{Frozen transfer to GraphCast and HRES forcing}
\begin{figure*}[!htbp]
  \centering
  \includegraphics[width=\textwidth]{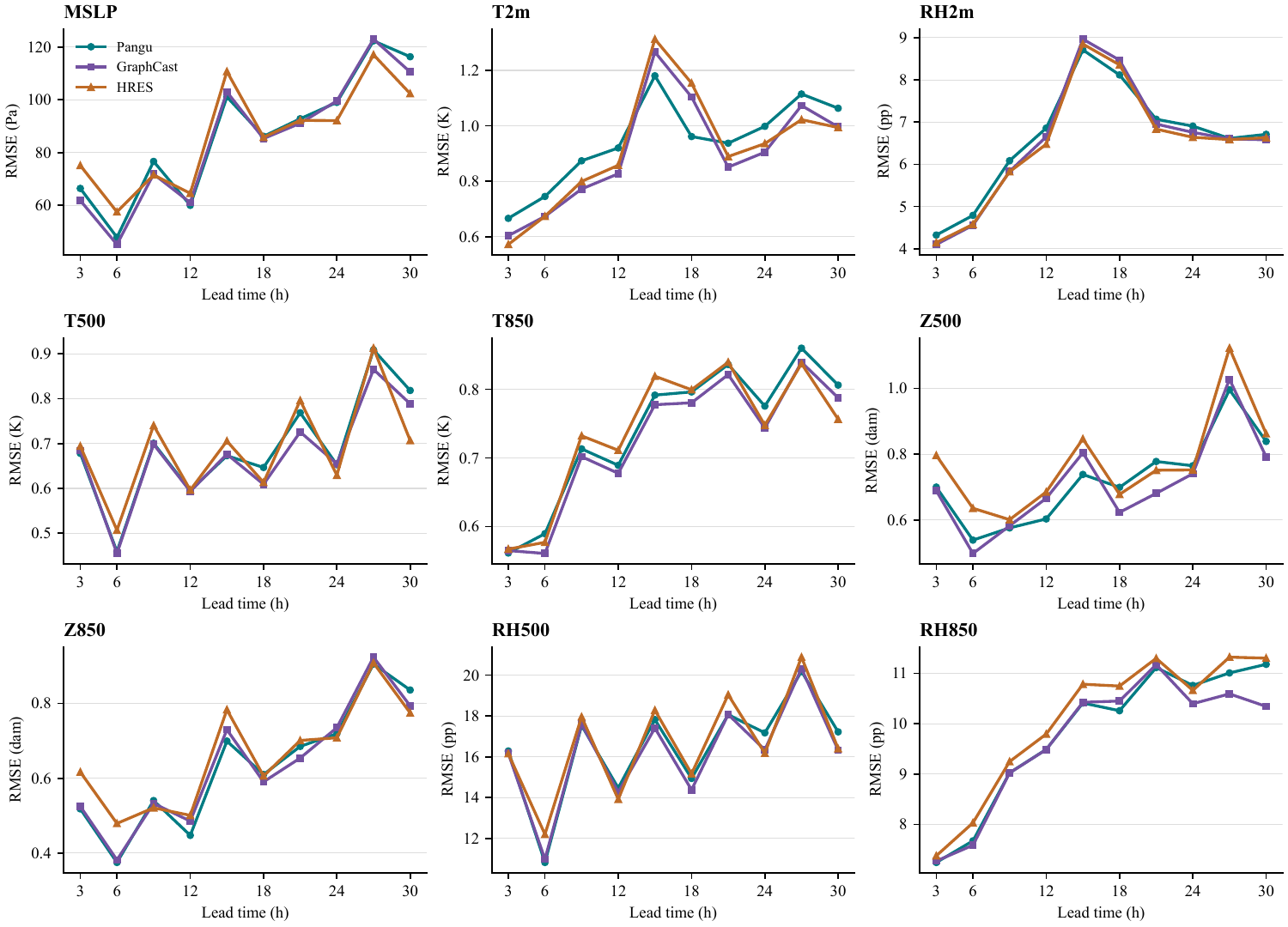}
  \caption{\textbf{Three-driver frozen-weight comparison.} Per-lead pixel-pooled RMSE for the same regional checkpoint driven by Pangu, GraphCast, or HRES.}
  \label{fig:global-forcing-transfer}
\end{figure*}

GraphCast and HRES reduce MSLP, T2m, and RH2m errors relative to Pangu-Weather, while upper-air performance varies by variable. The comparison measures driver substitution without regional retraining, and Figure~\ref{fig:global-forcing-transfer} shows how these differences evolve with forecast lead time.

\subsubsection{Frozen transfer from Pangu to GFS forcing}
\begin{figure*}[!htbp]
  \centering
  \includegraphics[width=\textwidth]{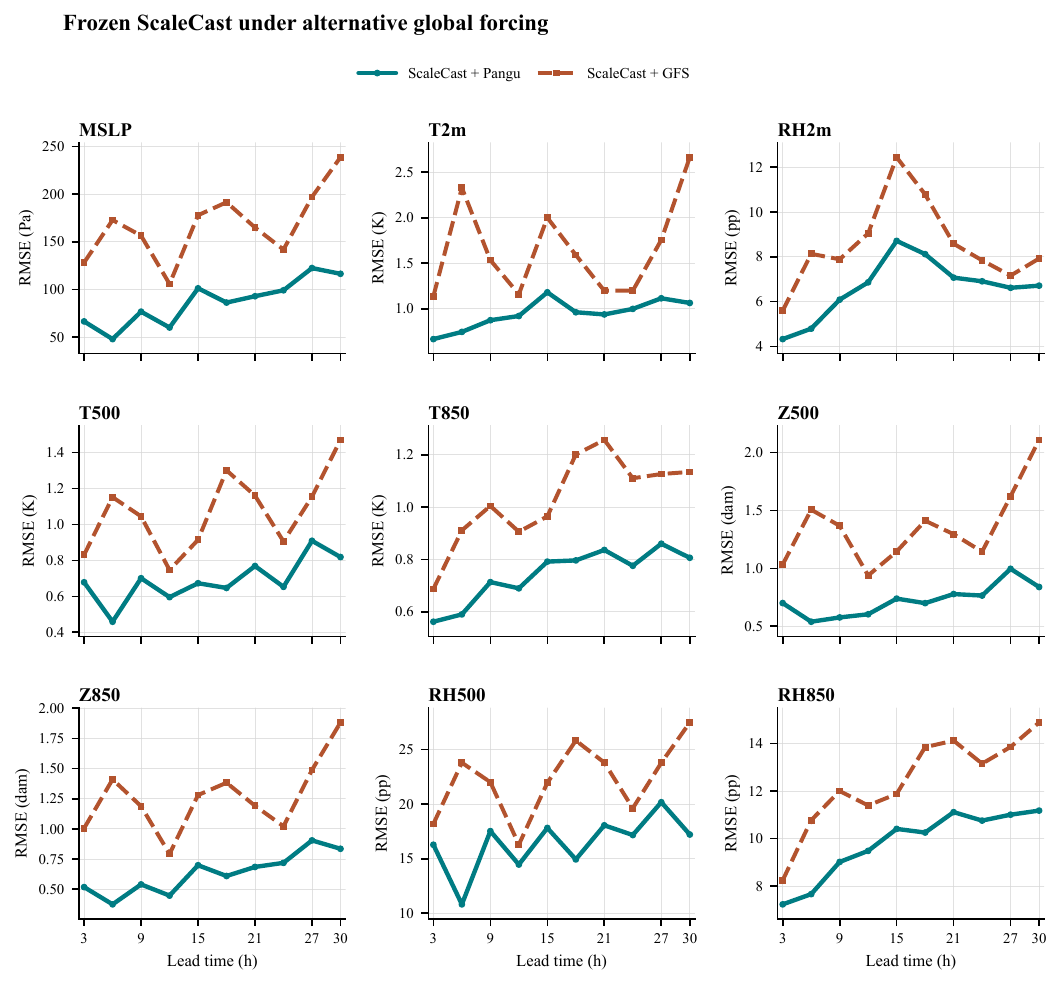}
  \caption{\textbf{Lead-dependent driver-shift test.} RMSE of the frozen ScaleCast checkpoint under fixed-initialization Pangu and GFS forcing. CERRA analyses are used only for verification.}
  \label{fig:v25-gfs-driver-30h}
\end{figure*}

We test whether the regional model can consume a global forecast source that was not used to produce its reported rollout results. The ScaleCast checkpoint, regional initialization, and ERA5-derived normalization are frozen, while the fixed-initialization Pangu trajectory is replaced by the corresponding 00~UTC GFS operational cycle. The archive contains the GFS $f000$ analysis and 12-hourly forecasts; these fields are linearly interpolated to the 3-hour regional-model steps. Table~\ref{tab:v25-gfs-driver-30h} shows that the frozen model remains numerically stable with GFS forcing but loses accuracy for all nine variables. The pooled increase ranges from 26.7\% for RH850 to 98.6\% for Z850. Figure~\ref{fig:v25-gfs-driver-30h} shows lead-dependent oscillation. Because the GFS archive has 12-hour rather than 6-hour native spacing, this test cannot separate driver distribution shift from temporal-interpolation effects. It demonstrates input compatibility, not driver invariance or the outcome of driver-specific retraining.

\definecolor{driverbestcell}{HTML}{DDF2EA}
\newcommand{\driverbest}[1]{\cellcolor{driverbestcell}\textbf{#1}}

\begin{table*}[!htbp]
  \centering
  \caption{\textbf{Frozen-driver transfer through $+30\,$h.} Pixel-pooled RMSE over ten 3-hour lead times. $\Delta$ is the relative change from Pangu to GFS forcing; lower is better.}
  \label{tab:v25-gfs-driver-30h}
  \setlength{\tabcolsep}{4.3pt}
  \renewcommand{\arraystretch}{1.08}
  \small
  \begin{tabular}{@{}llrrr@{}}
    \toprule
    Variable & Unit & ScaleCast + Pangu $\downarrow$ & ScaleCast + GFS $\downarrow$ & $\Delta$ (\%) \\
    \midrule
      MSLP & Pa & \driverbest{89.9} & 171.3 & +90.6 \\
      T2m & K & \driverbest{0.958} & 1.732 & +80.9 \\
      RH2m & pp & \driverbest{6.737} & 8.727 & +29.5 \\
      T500 & K & \driverbest{0.700} & 1.088 & +55.5 \\
      T850 & K & \driverbest{0.748} & 1.043 & +39.4 \\
      Z500 & dam & \driverbest{0.735} & 1.395 & +89.9 \\
      Z850 & dam & \driverbest{0.653} & 1.297 & +98.6 \\
      RH500 & pp & \driverbest{16.629} & 22.520 & +35.4 \\
      RH850 & pp & \driverbest{9.908} & 12.553 & +26.7 \\
    \bottomrule
  \end{tabular}
  \vspace{2pt}
\end{table*}


\subsubsection{Complete 30-hour lead curves}
\begin{figure*}[!htbp]
  \centering
  \includegraphics[width=\textwidth]{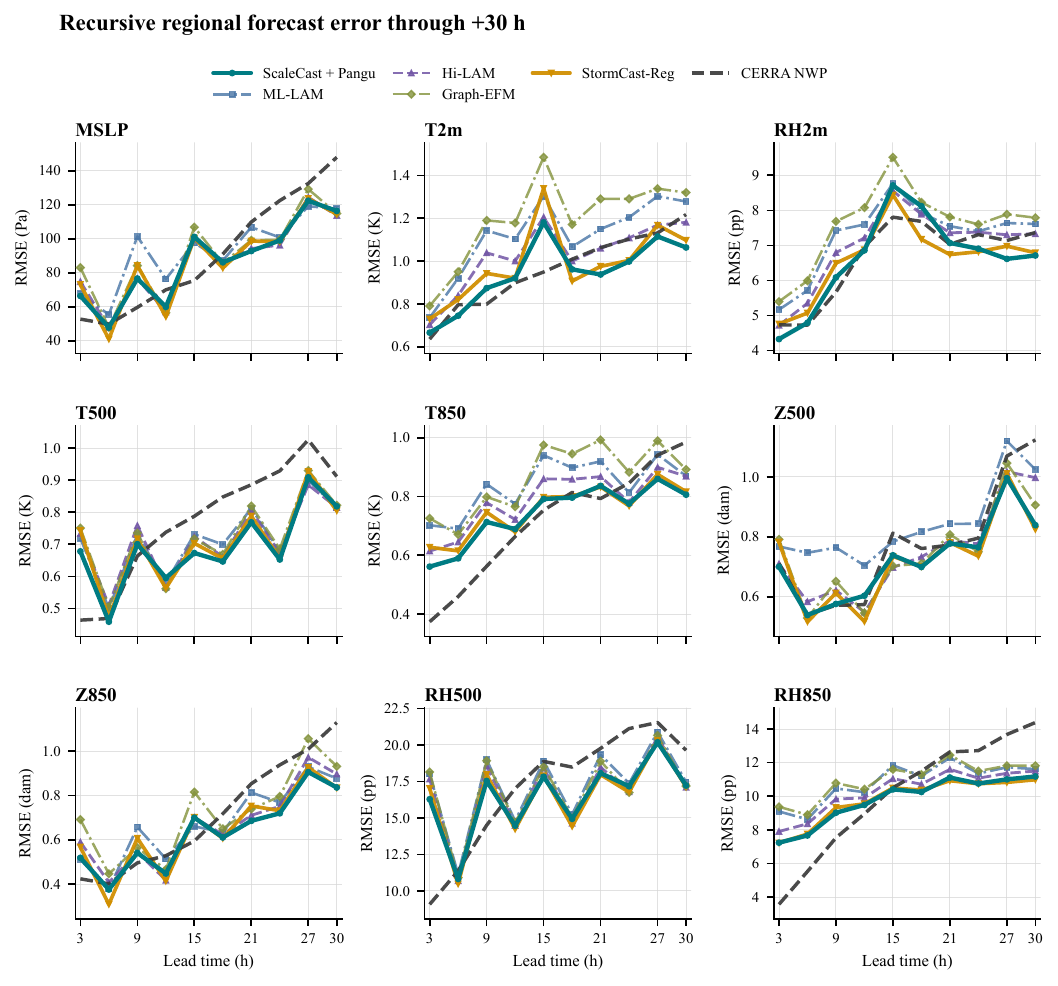}
  \caption{\textbf{Complete recursive forecast trajectories through $+30\,$h.} Pixel-pooled RMSE for the completed regional forecasting models, together with the same-cycle CERRA numerical forecast.}
  \label{fig:all-models-30h}
\end{figure*}

Figure~\ref{fig:all-models-30h} reports the shared-Pangu regional-model subset and CERRA NWP from Table~\ref{tab:v25-v33-cerra-30h}. ScaleCast, Hi-LAM, and StormCast-Reg remain comparatively stable across the ten recursive steps. The humidity variables show substantial case-dependent oscillation, so pooled scalar scores should be interpreted together with these trajectories.

\begin{figure*}[!htbp]
  \centering
  \includegraphics[width=\textwidth]{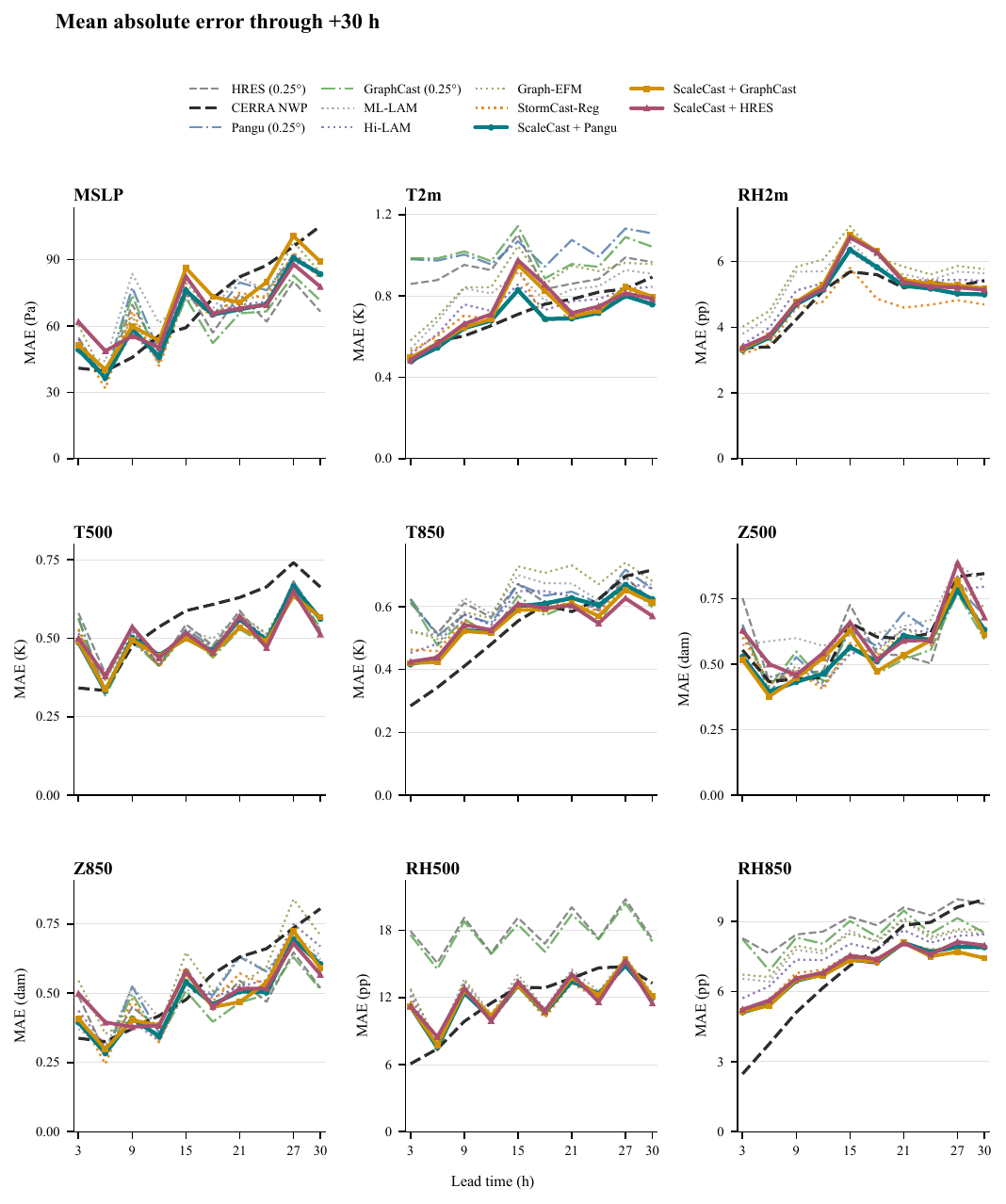}
  \caption{\textbf{Mean absolute error through $+30\,$h.} Per-lead pixel-pooled MAE for the methods in Table~\ref{tab:v25-v33-cerra-30h}. Solid curves highlight the frozen ScaleCast driver variants; dashed and dotted curves show external, global, and regional comparison systems. Missing curves denote diagnostics unavailable from the source forecast.}
  \label{fig:table2-mae-by-lead}
\end{figure*}

Table~\ref{tab:v25-v33-cerra-30h} summarizes squared-error sensitivity through pooled RMSE. Figures~\ref{fig:table2-mae-by-lead} report the corresponding per-lead mean absolute error (MAE). MAE reduces the influence of isolated large departures, while bias reveals directional drift that RMSE cannot identify. 

\subsubsection{Absolute regional forecast fields}
\label{app:regional-fields}
\begin{figure}[!htbp]
\centering
\includegraphics[width=\linewidth]{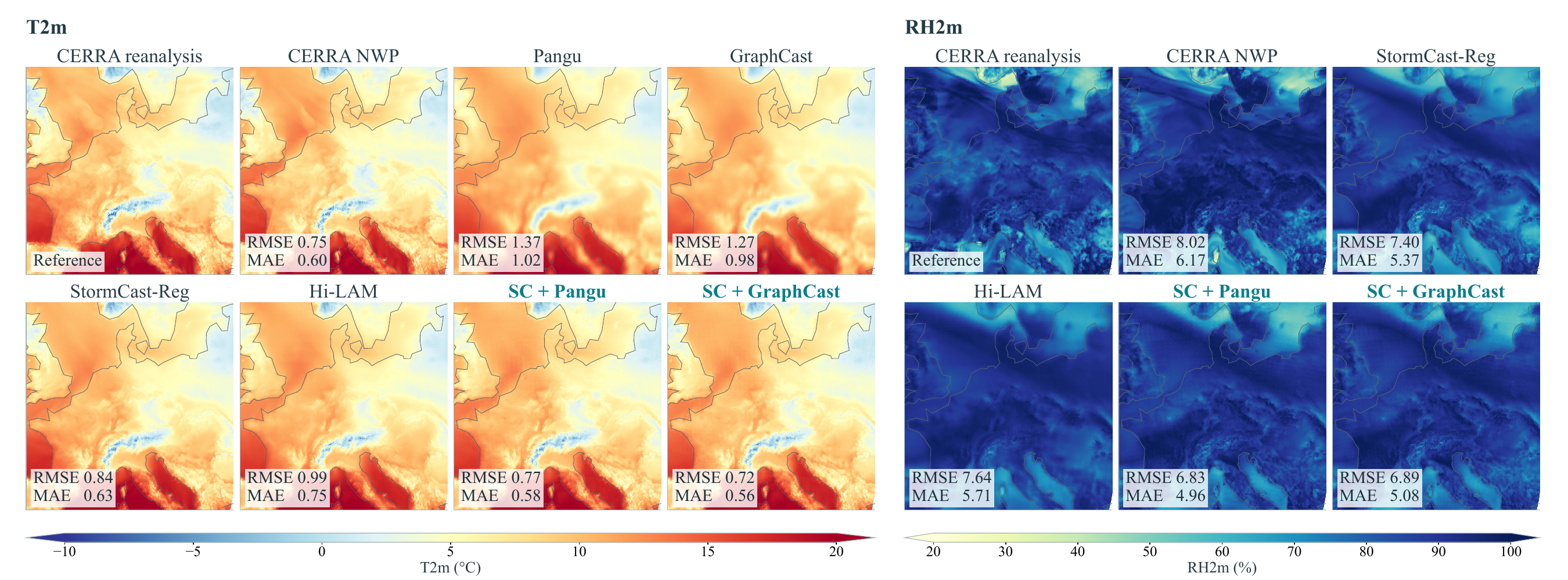}
\caption{Absolute fields for two cases: T2m from 16 November 2022, 00 UTC at $+9\,$h (left), and RH2m from 5 April 2022, 00 UTC at $+27\,$h (right). Each variable uses a shared colour scale across the reference and forecasts, in $^\circ$C and \%, respectively. The square window and post-hoc case selection match Figure~\ref{fig:regional-error}.}
\label{fig:regional-absolute}
\end{figure}

Figure~\ref{fig:regional-absolute} provides the absolute fields underlying Figure~\ref{fig:regional-error}. 

\FloatBarrier
\subsection{Weather-system and station case studies}

\label{app:windstorm-candidates}
\begin{figure*}[!htbp]
  \centering
  \includegraphics[width=\textwidth]{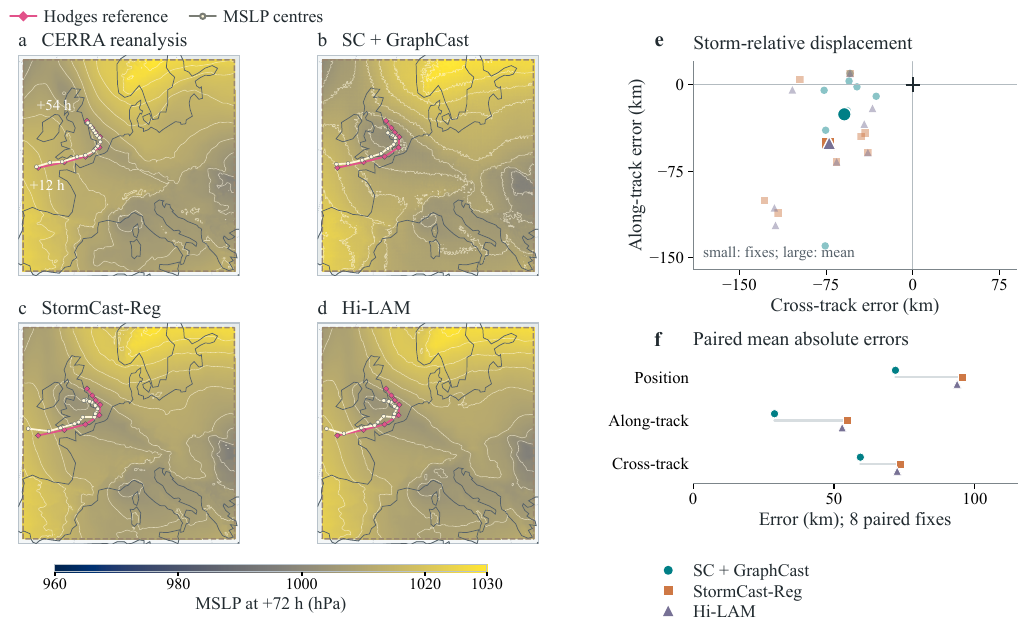}
  \caption{\textbf{November case and storm-relative displacement.} Event 1812, initialized at 00 UTC on 16 November 2022. \textbf{a--d,} MSLP at $+72\,$h and cumulative Hodges and independently tracked field centres. \textbf{e,} Along- and cross-track displacement at eight shared fixes with equal kilometre scales; positive values indicate ahead of and right of reference motion. Large symbols mark means. \textbf{f,} Mean distance and mean absolute displacement components.}
  \label{fig:windstorm-november}
\end{figure*}

\subsubsection{Windstorm tracking across the candidate events}

\begin{figure*}[!htbp]
  \centering
  \includegraphics[page=1,trim=0 0 0 11bp,clip,width=\textwidth]{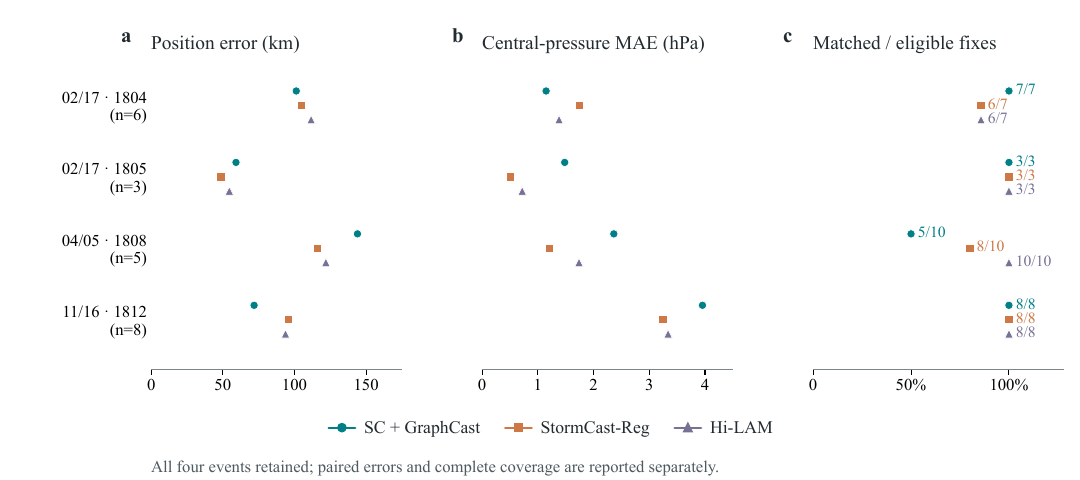}
  \caption{\textbf{Archived four-event diagnostics.} Paired errors use shared times ($n$); coverage uses all eligible fixes. Current event-1804 position results appear in Figure~\ref{fig:windstorm-selected}.}
  \label{fig:windstorm-complete-summary}
\end{figure*}

\begin{figure*}[!htbp]
  \centering
  \includegraphics[page=2,trim=0 14bp 0 12bp,clip,width=\textwidth]{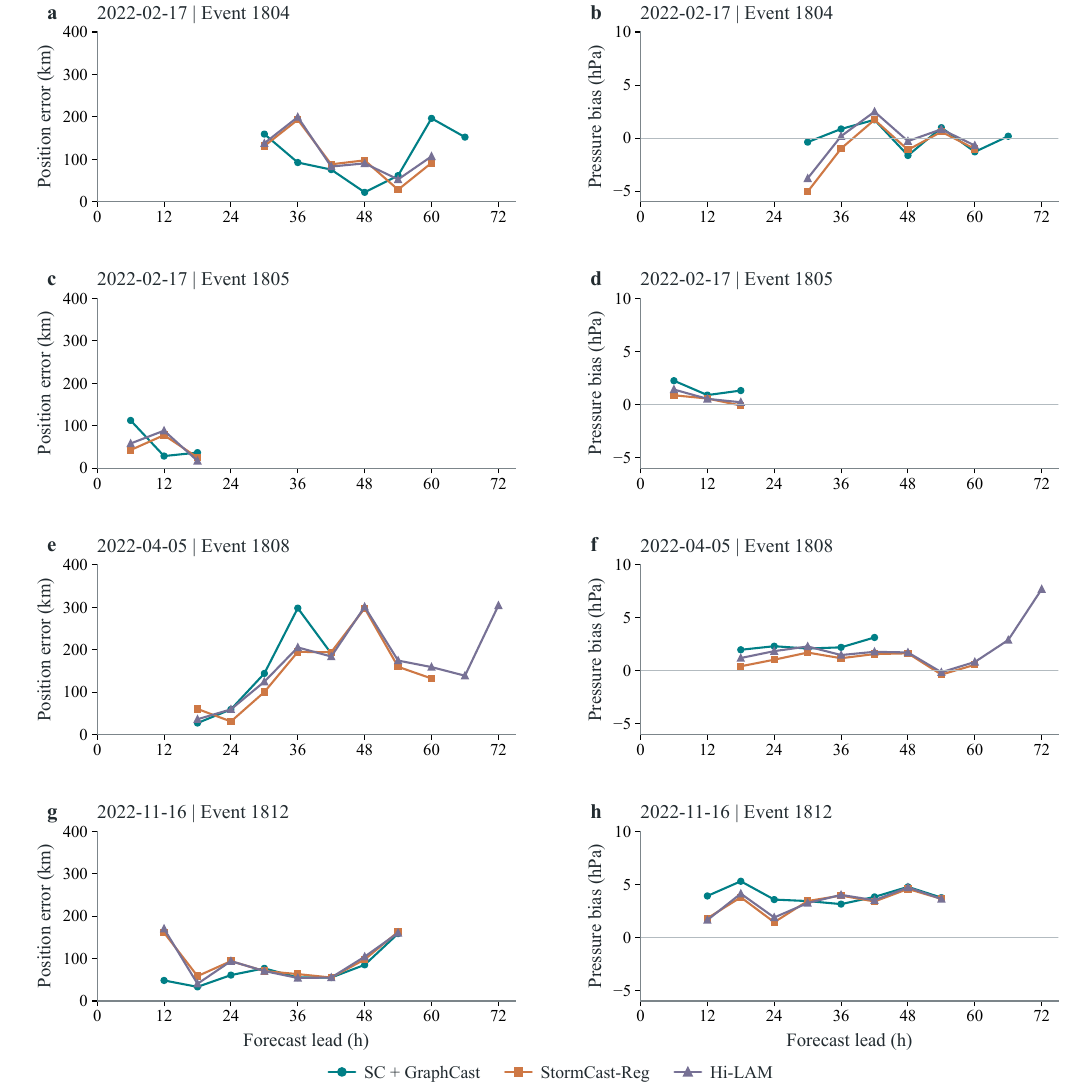}
  \caption{\textbf{Archived four-event lead curves.} Left: position error. Right: signed central-pressure error. Current event-1804 position results appear in Figure~\ref{fig:windstorm-selected}.}
  \label{fig:windstorm-complete-leads}
\end{figure*}

\begin{figure*}[!htbp]
  \centering
  \includegraphics[width=\textwidth]{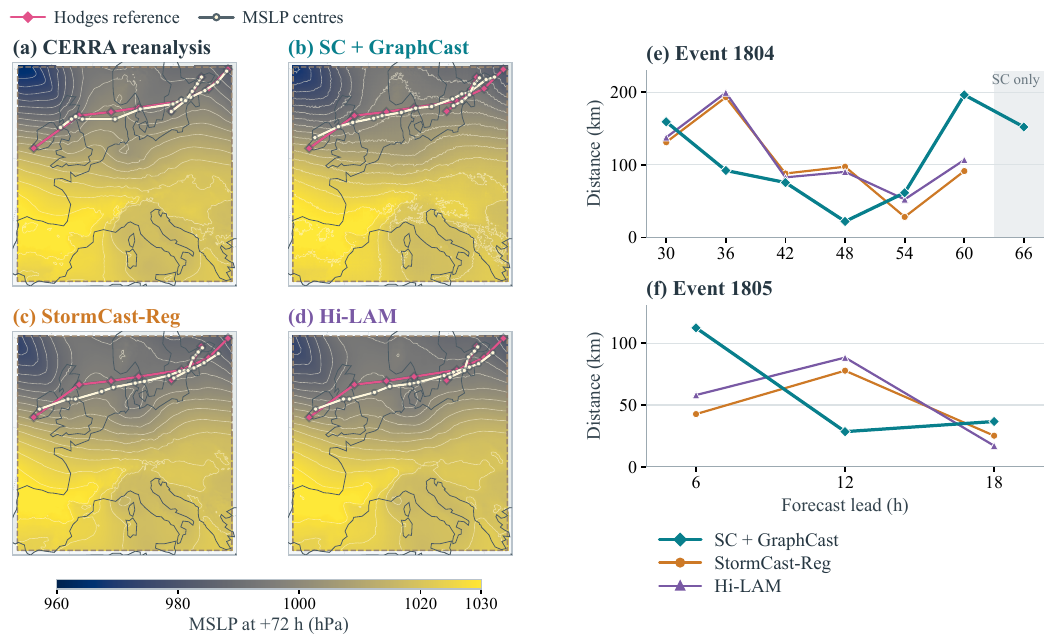}
  \caption{\textbf{Archived February tracking diagnostics.} Initialization: 17 February 2022, 00 UTC. Maps show MSLP at $+72\,$h and cumulative Hodges and field-derived tracks; the right panels show position errors for events 1804 and 1805. The event-1804 curve predates the current comparison in Figure~\ref{fig:windstorm-selected}. The shaded $+66\,$h point is matched only by SC + GraphCast.}
  \label{fig:windstorm-february}
\end{figure*}

Figure~\ref{fig:windstorm-selected} reports the current event-1804 position and core-pressure comparison. Figures~\ref{fig:windstorm-complete-summary}--\ref{fig:windstorm-february} retain archived diagnostics from four events in the forecast windows of 17 February, 5 April, and 16 November 2022 (00 UTC). Their event-1804 position values precede the main-text revision and are not used for its reported comparison. Figure~\ref{fig:windstorm-november} supplements the main case with event 1812 and its along- and cross-track displacement.

\FloatBarrier

\subsubsection{Spatial station-error diagnostics}
\label{app:station-spatial}
\begin{figure*}[!htbp]
  \centering
  \includegraphics[width=\textwidth]{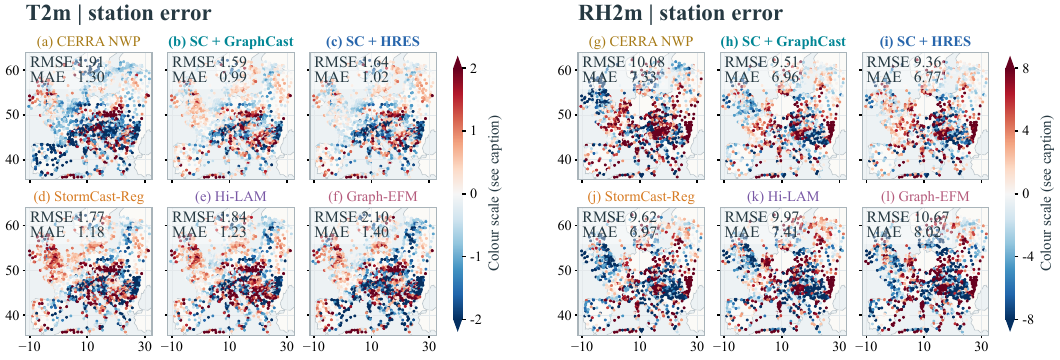}
  \caption{\textbf{Spatial temperature and humidity errors against HadISD.}
  Initialization: 17 February 2022, 00 UTC; lead: $+9\,$h.
  \textbf{a--f,} T2m errors at 1,626 common stations; \textbf{g--l,} RH2m errors at 1,140 common stations. Colours show forecast minus observed values.}
  \label{fig:station-spatial-errors}
\end{figure*}

Figure~\ref{fig:station-spatial-errors} retains the spatial panels accompanying the station trajectories in Figure~\ref{fig:station-t2m-rh2m}. All methods use the same valid station mask within each variable.

\FloatBarrier
\subsection{Design and training comparisons}

The following one-step studies examine whether training objectives and the geographical extent of the coarse input change validation error. They use target-time global analyses, whereas the main recursive comparison uses forecast trajectories. Their scores therefore support design diagnostics, not a direct ranking against the $+30$-hour results.

\paragraph{Main-table replacement settings.}
\label{app:module-settings}
In Table~\ref{tab:module-replacement-ablation}, cross-attention replaces regional attention, not GRConv. Geo-remap + fusion and GRConv use the same regional backbone, training budget, and loss. Direct prediction removes only the increment output, while Shared and Separate differ only in their output heads, with the optimization unchanged.

\subsubsection{Training strategies and humidity objectives}
\label{app:training-comparisons}

\begin{table*}[!htbp]
  \centering
  \setlength{\belowcaptionskip}{6pt}
  \caption{\textbf{Additional training comparisons.} RMSE and normalized MSE on 2,919 validation samples from 2023 at $+3\,$h, conditioned on target-time ERA5 analyses. Lower is better. Bold and underlined values indicate the best and second-best results within each block.}
  \label{tab:training-comparisons}
  \begin{threeparttable}
    \setlength{\tabcolsep}{3pt}
    \renewcommand{\arraystretch}{1.18}
    \scriptsize
    \resizebox{\textwidth}{!}{%
    \begin{tabular}{@{}llrrrrrrrrrr@{}}
      \toprule
      \multicolumn{2}{c}{Design choices} & \multicolumn{3}{c}{Surface RMSE} & \multicolumn{6}{c}{Upper-air RMSE} & \\
      \cmidrule(lr){1-2}\cmidrule(lr){3-5}\cmidrule(lr){6-11}
      Module & Training & \shortstack{MSLP $\downarrow$\\(Pa)} & \shortstack{T2m $\downarrow$\\(K)} & \shortstack{RH2m $\downarrow$\\(pp)} & \shortstack{T500 $\downarrow$\\(K)} & \shortstack{T850 $\downarrow$\\(K)} & \shortstack{Z500 $\downarrow$\\(dam)} & \shortstack{Z850 $\downarrow$\\(dam)} & \shortstack{RH500 $\downarrow$\\(pp)} & \shortstack{RH850 $\downarrow$\\(pp)} & \shortstack{nMSE $\downarrow$\\($10^{-2}$)} \\
      \midrule
      \multicolumn{12}{@{}l}{\textit{Spatial training strategies}} \\
      \addlinespace[2pt]
      Scale + terrain & Grouped & \abest{40.8} & \abest{0.766} & \asecond{5.336} & \abest{0.383} & \abest{0.462} & \abest{0.476} & \abest{0.336} & \abest{8.173} & \abest{6.217} & \abest{3.009} \\
      Scale + terrain & MS + grad. & \asecond{43.4} & \asecond{0.779} & 5.366 & \asecond{0.383} & \asecond{0.464} & \asecond{0.500} & \asecond{0.357} & \asecond{8.189} & \asecond{6.225} & \asecond{3.038} \\
      Scale + terrain$^{10}$ & FAMO + noise & 55.8 & 0.782 & \abest{5.253} & 0.428 & 0.489 & 0.626 & 0.469 & 9.009 & 6.644 & 3.305 \\
      \midrule
      \multicolumn{12}{@{}l}{\textit{Humidity objectives}} \\
      \addlinespace[2pt]
      Low-rank + res. & FAMO & \abest{39.3} & \asecond{0.723} & \asecond{5.283} & \abest{0.374} & \abest{0.450} & \abest{0.450} & \abest{0.328} & \asecond{8.196} & \abest{6.228} & \asecond{2.974} \\
      Low-rank + res.$^{14}$ & FAMO + RH & \asecond{41.1} & \abest{0.722} & \abest{5.176} & \asecond{0.395} & \asecond{0.464} & \asecond{0.471} & \asecond{0.337} & \abest{8.121} & \asecond{6.239} & \abest{2.922} \\
      \bottomrule
    \end{tabular}%
    }
  \end{threeparttable}
\end{table*}


Table~\ref{tab:training-comparisons} compares selected configurations with a $128\times128$ coarse input on 2,919 validation samples from 2023 at $+3\,$h. Each is conditioned on target-time ERA5 analyses.

\paragraph{Spatial training strategies.}
With the scale-and-terrain configuration, multiscale and gradient losses do not improve aggregate nMSE over grouped optimization: the displayed value rises from $3.009$ to $3.038$ in units of $10^{-2}$. The FAMO configuration with coarse-field perturbations lowers RH2m RMSE from $5.336$ to $5.253$ percentage points, but MSLP and Z500 errors rise from $40.8$ to $55.8$~Pa and $0.476$ to $0.626$~dam. The local humidity gain is therefore accompanied by losses in other fields, rather than an across-variable improvement.

\paragraph{Humidity objectives.}
Adding humidity objectives to FAMO lowers RH2m and RH500 RMSE from $5.283$ to $5.176$ and from $8.196$ to $8.121$ percentage points. Aggregate nMSE falls from $2.974$ to $2.922$ in units of $10^{-2}$, while MSLP, upper-air temperature, and geopotential errors increase. These rows illustrate a multi-variable trade-off at the selected checkpoints. Training budgets differ, so they cannot isolate the causal effect of the objective alone.

\subsubsection{Extent of the global input}
\label{app:input-coverage}
\begin{figure}[!htbp]
  \centering
  \includegraphics[width=\linewidth]{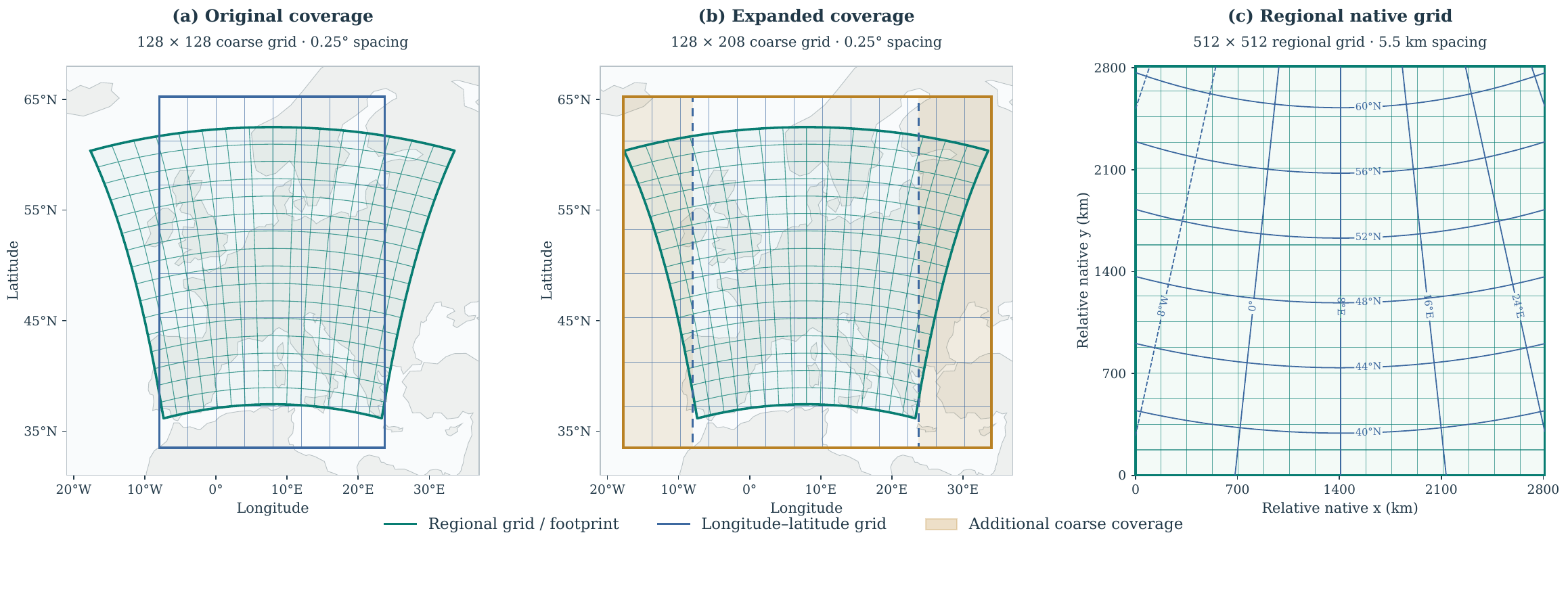}
  \caption{\textbf{Coarse and regional grids in different coordinate systems.} (a,b) Actual footprints of the $128\times128$ and $128\times208$ coarse grids, with $0.25^\circ$ angular spacing, overlaid on the $512\times512$ CERRA grid. Expansion covers the regional side areas outside the original coarse bounds; the dashed outline marks the original footprint. (c) The same regional grid in relative native coordinates, with approximately 5.5~km spacing and geographic coordinate contours. Grid lines are subsampled.}
  \label{fig:input-coverage}
\end{figure}

We compare $128\times128$ and $128\times208$ coarse inputs while holding the $512\times512$ regional grid fixed. Figure~\ref{fig:input-coverage} shows that the smaller coarse domain already contains the entire regional domain; expansion adds surrounding atmospheric context rather than changing the predicted area or its resolution.

\begin{thebibliography}{45}
\providecommand{\natexlab}[1]{#1}
\providecommand{\url}[1]{\texttt{#1}}
\expandafter\ifx\csname urlstyle\endcsname\relax
  \providecommand{\doi}[1]{doi: #1}\else
  \providecommand{\doi}{doi: \begingroup \urlstyle{rm}\Url}\fi

\bibitem[Abdi et~al.(2026)Abdi, Jankov, Madden, Vargas, Smith, Frolov, Flora,
  and Potvin]{abdi2026hrrrcast}
Daniel Abdi, Isidora Jankov, Paul Madden, Vanderlei Vargas, Timothy~A Smith,
  Sergey Frolov, Montgomery Flora, and Corey Potvin.
\newblock Hrrrcast: A data-driven emulator for regional weather forecasting at
  convection-allowing scales.
\newblock \emph{Artificial Intelligence for the Earth Systems}, 5\penalty0
  (2):\penalty0 250061, 2026.

\bibitem[Adamov et~al.(2025)Adamov, Oskarsson, Denby, Landelius, Hintz,
  Christiansen, Schicker, Osuna, Lindsten, Fuhrer, et~al.]{adamov2025limited}
Simon Adamov, Joel Oskarsson, Leif Denby, Tomas Landelius, Kasper Hintz, Simon
  Christiansen, Irene Schicker, Carlos Osuna, Fredrik Lindsten, Oliver Fuhrer,
  et~al.
\newblock Building machine learning limited area models: Kilometer-scale
  weather forecasting in realistic settings.
\newblock \emph{arXiv preprint arXiv:2504.09340}, 2025.

\bibitem[Ba{\~n}o-Medina et~al.(2022)Ba{\~n}o-Medina, Manzanas, Cimadevilla,
  Fern{\'a}ndez, Gonz{\'a}lez-Abad, Cofi{\~n}o, and
  Guti{\'e}rrez]{banomedina2022deepesd}
Jorge Ba{\~n}o-Medina, Rodrigo Manzanas, Ezequiel Cimadevilla, Jes{\'u}s
  Fern{\'a}ndez, Jose Gonz{\'a}lez-Abad, Antonio~Santiago Cofi{\~n}o, and
  Jos{\'e}~Manuel Guti{\'e}rrez.
\newblock Downscaling multi-model climate projection ensembles with deep
  learning (deepesd): contribution to cordex eur-44.
\newblock \emph{Geoscientific Model Development Discussions}, 2022:\penalty0
  1--14, 2022.

\bibitem[Bauer et~al.(2015)Bauer, Thorpe, and Brunet]{bauer2015quiet}
Peter Bauer, Alan Thorpe, and Gilbert Brunet.
\newblock The quiet revolution of numerical weather prediction.
\newblock \emph{Nature}, 525\penalty0 (7567):\penalty0 47--55, 2015.

\bibitem[Bi et~al.(2023)Bi, Xie, Zhang, Chen, Gu, and Tian]{bi2023pangu}
Kaifeng Bi, Lingxi Xie, Hengheng Zhang, Xin Chen, Xiaotao Gu, and Qi~Tian.
\newblock Accurate medium-range global weather forecasting with 3d neural
  networks.
\newblock \emph{Nature}, 619\penalty0 (7970):\penalty0 533--538, 2023.

\bibitem[Bodnar et~al.(2025)Bodnar, Bruinsma, Lucic, Stanley, Allen,
  Brandstetter, Garvan, Riechert, Weyn, Dong, et~al.]{bodnar2025aurora}
Cristian Bodnar, Wessel~P Bruinsma, Ana Lucic, Megan Stanley, Anna Allen,
  Johannes Brandstetter, Patrick Garvan, Maik Riechert, Jonathan~A Weyn, Haiyu
  Dong, et~al.
\newblock A foundation model for the earth system.
\newblock \emph{Nature}, 641\penalty0 (8065):\penalty0 1180--1187, 2025.

\bibitem[Chen et~al.(2026)Chen, Han, Zhang, Guo, and Bai]{chen2026stcast}
Hao Chen, Tao Han, Jie Zhang, Song Guo, and Lei Bai.
\newblock Stcast: Adaptive boundary alignment for global and regional weather
  forecasting.
\newblock In \emph{Proceedings of the IEEE/CVF conference on computer vision
  and pattern recognition}, pp.\  20586--20596, 2026.

\bibitem[Chen et~al.(2025)Chen, Han, Ling, Gong, Bai, Wang, Luo, Fei, Zhang,
  Chen, et~al.]{chen2025operational}
Kang Chen, Tao Han, Fenghua Ling, Junchao Gong, Lei Bai, Xinyu Wang, Jing-Jia
  Luo, Ben Fei, Wenlong Zhang, Xi~Chen, et~al.
\newblock The operational medium-range deterministic weather forecasting can be
  extended beyond a 10-day lead time.
\newblock \emph{Communications Earth \& Environment}, 6\penalty0 (1):\penalty0
  518, 2025.

\bibitem[Chen et~al.(2023)Chen, Zhong, Zhang, Cheng, Xu, Qi, and
  Li]{chen2023fuxi}
Lei Chen, Xiaohui Zhong, Feng Zhang, Yuan Cheng, Yinghui Xu, Yuan Qi, and Hao
  Li.
\newblock Fuxi: A cascade machine learning forecasting system for 15-day global
  weather forecast.
\newblock \emph{npj climate and atmospheric science}, 6\penalty0 (1):\penalty0
  190, 2023.

\bibitem[Clark et~al.(2016)Clark, Roberts, Lean, Ballard, and
  Charlton-Perez]{clark2016convection}
Peter Clark, Nigel Roberts, Humphrey Lean, Susan~P Ballard, and Cristina
  Charlton-Perez.
\newblock Convection-permitting models: A step-change in rainfall forecasting.
\newblock \emph{Meteorological Applications}, 23\penalty0 (2):\penalty0
  165--181, 2016.

\bibitem[{Copernicus Climate Change Service}(2025)]{c3s2025windstorm}
{Copernicus Climate Change Service}.
\newblock Windstorm tracks and footprints derived from reanalysis over europe
  between 1940 to present, 2025.

\bibitem[Dowell et~al.(2022)Dowell, Alexander, James, Weygandt, Benjamin,
  Manikin, Blake, Brown, Olson, Hu, et~al.]{dowell2022hrrr}
David~C Dowell, Curtis~R Alexander, Eric~P James, Stephen~S Weygandt, Stanley~G
  Benjamin, Geoffrey~S Manikin, Benjamin~T Blake, John~M Brown, Joseph~B Olson,
  Ming Hu, et~al.
\newblock The high-resolution rapid refresh (hrrr): An hourly updating
  convection-allowing forecast model. part i: Motivation and system
  description.
\newblock \emph{Weather and Forecasting}, 37\penalty0 (8):\penalty0 1371--1395,
  2022.

\bibitem[Dunn et~al.(2012)Dunn, Willett, Thorne, Woolley, Durre, Dai, Parker,
  and Vose]{dunn2012hadisd}
Robert~JH Dunn, Kate~M Willett, Peter~W Thorne, Emma~V Woolley, Imke Durre,
  Aiguo Dai, David~E Parker, and RS~Vose.
\newblock Hadisd: A quality-controlled global synoptic report database for
  selected variables at long-term stations from 1973--2011.
\newblock \emph{Climate of the Past}, 8\penalty0 (5):\penalty0 1649--1679,
  2012.

\bibitem[Dunn et~al.(2016)Dunn, Willett, Parker, and
  Mitchell]{dunn2016expanding}
Robert~JH Dunn, Kate~M Willett, David~E Parker, and Lorna Mitchell.
\newblock Expanding hadisd: Quality-controlled, sub-daily station data from
  1931.
\newblock \emph{Geoscientific Instrumentation, Methods and Data Systems},
  5\penalty0 (2):\penalty0 473--491, 2016.

\bibitem[Gao et~al.(2025)Gao, Wu, Shu, Dong, Xu, Chen, Yan, Wen, Hu, Wang,
  et~al.]{gao2025oneforecast}
Yuan Gao, Hao Wu, Ruiqi Shu, Huanshuo Dong, Fan Xu, Rui~Ray Chen, Yibo Yan,
  Qingsong Wen, Xuming Hu, Kun Wang, et~al.
\newblock Oneforecast: a universal framework for global and regional weather
  forecasting.
\newblock In \emph{Proceedings of the 42nd International Conference on Machine
  Learning}, pp.\  18658--18697, 2025.

\bibitem[Harris et~al.(2022)Harris, McRae, Chantry, Dueben, and
  Palmer]{harris2022stochastic}
Lucy Harris, Andrew~TT McRae, Matthew Chantry, Peter~D Dueben, and Tim~N
  Palmer.
\newblock A generative deep learning approach to stochastic downscaling of
  precipitation forecasts.
\newblock \emph{Journal of Advances in Modeling Earth Systems}, 14\penalty0
  (10):\penalty0 e2022MS003120, 2022.

\bibitem[He et~al.(2016)He, Zhang, Ren, and Sun]{he2016resnet}
Kaiming He, Xiangyu Zhang, Shaoqing Ren, and Jian Sun.
\newblock Deep residual learning for image recognition.
\newblock In \emph{Proceedings of the IEEE conference on computer vision and
  pattern recognition}, pp.\  770--778, 2016.

\bibitem[Hersbach et~al.(2020)Hersbach, Bell, Berrisford, Hirahara,
  Hor{\'a}nyi, Mu{\~n}oz-Sabater, Nicolas, Peubey, Radu, Schepers,
  et~al.]{hersbach2020era5}
Hans Hersbach, Bill Bell, Paul Berrisford, Shoji Hirahara, Andr{\'a}s
  Hor{\'a}nyi, Joaqu{\'\i}n Mu{\~n}oz-Sabater, Julien Nicolas, Carole Peubey,
  Raluca Radu, Dinand Schepers, et~al.
\newblock The era5 global reanalysis.
\newblock \emph{Quarterly journal of the royal meteorological society},
  146\penalty0 (730):\penalty0 1999--2049, 2020.

\bibitem[H{\"o}hlein et~al.(2024)H{\"o}hlein, Schulz, Westermann, and
  Lerch]{hohlein2024postprocessing}
Kevin H{\"o}hlein, Benedikt Schulz, R{\"u}diger Westermann, and Sebastian
  Lerch.
\newblock Postprocessing of ensemble weather forecasts using
  permutation-invariant neural networks.
\newblock \emph{Artificial Intelligence for the Earth Systems}, 3\penalty0
  (1):\penalty0 e230070, 2024.

\bibitem[Keisler(2022)]{keisler2022forecasting}
Ryan Keisler.
\newblock Forecasting global weather with graph neural networks.
\newblock \emph{arXiv preprint arXiv:2202.07575}, 2022.

\bibitem[Kochkov et~al.(2024)Kochkov, Yuval, Langmore, Norgaard, Smith, Mooers,
  Kl{\"o}wer, Lottes, Rasp, D{\"u}ben, et~al.]{kochkov2024neural}
Dmitrii Kochkov, Janni Yuval, Ian Langmore, Peter Norgaard, Jamie Smith,
  Griffin Mooers, Milan Kl{\"o}wer, James Lottes, Stephan Rasp, Peter
  D{\"u}ben, et~al.
\newblock Neural general circulation models for weather and climate.
\newblock \emph{Nature}, 632\penalty0 (8027):\penalty0 1060--1066, 2024.

\bibitem[Laitenberger et~al.(2026)Laitenberger, Kopiczko, Snoek, and
  Asano]{laitenberger2026gateskip}
Filipe Laitenberger, Dawid Kopiczko, Cees~G Snoek, and Yuki Asano.
\newblock What layers when: Learning to skip compute in llms with residual
  gates.
\newblock In \emph{International Conference on Learning Representations},
  volume 2026, pp.\  78161--78184, 2026.

\bibitem[Lam et~al.(2023)Lam, Sanchez-Gonzalez, Willson, Wirnsberger,
  Fortunato, Alet, Ravuri, Ewalds, Eaton-Rosen, Hu, et~al.]{lam2023graphcast}
Remi Lam, Alvaro Sanchez-Gonzalez, Matthew Willson, Peter Wirnsberger, Meire
  Fortunato, Ferran Alet, Suman Ravuri, Timo Ewalds, Zach Eaton-Rosen, Weihua
  Hu, et~al.
\newblock Learning skillful medium-range global weather forecasting.
\newblock \emph{Science}, 382\penalty0 (6677):\penalty0 1416--1421, 2023.

\bibitem[Lang et~al.(2024)Lang, Alexe, Chantry, Dramsch, Pinault, Raoult,
  Clare, Lessig, Maier-Gerber, Magnusson, et~al.]{lang2024aifs}
Simon Lang, Mihai Alexe, Matthew Chantry, Jesper Dramsch, Florian Pinault,
  Baudouin Raoult, Mariana~CA Clare, Christian Lessig, Michael Maier-Gerber,
  Linus Magnusson, et~al.
\newblock Aifs--ecmwf's data-driven forecasting system.
\newblock \emph{arXiv preprint arXiv:2406.01465}, 2024.

\bibitem[Larsson et~al.(2025)Larsson, Oskarsson, Landelius, and
  Lindsten]{larsson2025diffusion}
Erik Larsson, Joel Oskarsson, Tomas Landelius, and Fredrik Lindsten.
\newblock Diffusion-lam: probabilistic limited area weather forecasting with
  diffusion.
\newblock \emph{arXiv preprint arXiv:2502.07532}, 2025.

\bibitem[Liu et~al.(2023)Liu, Feng, Stone, and Liu]{liu2023famo}
Bo~Liu, Yihao Feng, Peter Stone, and Qiang Liu.
\newblock Famo: Fast adaptive multitask optimization.
\newblock \emph{Advances in Neural Information Processing Systems},
  36:\penalty0 57226--57243, 2023.

\bibitem[Mardani et~al.(2025)Mardani, Brenowitz, Cohen, Pathak, Chen, Liu,
  Vahdat, Nabian, Ge, Subramaniam, et~al.]{mardani2025corrdiff}
Morteza Mardani, Noah Brenowitz, Yair Cohen, Jaideep Pathak, Chieh-Yu Chen,
  Cheng-Chin Liu, Arash Vahdat, Mohammad~Amin Nabian, Tao Ge, Akshay
  Subramaniam, et~al.
\newblock Residual corrective diffusion modeling for km-scale atmospheric
  downscaling.
\newblock \emph{Communications Earth \& Environment}, 6\penalty0 (1):\penalty0
  124, 2025.

\bibitem[Merizzi et~al.(2024)Merizzi, Asperti, and Colamonaco]{merizzi2024wind}
Fabio Merizzi, Andrea Asperti, and Stefano Colamonaco.
\newblock Wind speed super-resolution and validation: from era5 to cerra via
  diffusion models.
\newblock \emph{Neural Computing and Applications}, 36\penalty0 (34):\penalty0
  21899--21921, 2024.

\bibitem[Molinaro et~al.(2026)Molinaro, Siegenheim, Martin, Frey, Poulsen,
  Seitz, and Gabler]{molinaro2026universal}
Roberto Molinaro, Niall Siegenheim, Henry Martin, Mark Frey, Niels Poulsen,
  Philipp Seitz, and Marvin~Vincent Gabler.
\newblock Universal diffusion-based probabilistic downscaling.
\newblock \emph{arXiv preprint arXiv:2602.11893}, 2026.

\bibitem[Mousavi et~al.(2026)Mousavi, Wen, Lingsch, Herde, Raonic, and
  Mishra]{mousavi2025rigno}
Sepehr Mousavi, Shizheng Wen, Levi Lingsch, Maximilian Herde, Bogdan Raonic,
  and Siddhartha Mishra.
\newblock Rigno: A graph-based framework for robust and accurate operator
  learning for pdes on arbitrary domains.
\newblock \emph{Advances in Neural Information Processing Systems},
  38:\penalty0 150039--150101, 2026.

\bibitem[Nipen et~al.(2026)Nipen, Haugen, Ingstad, Nordhagen, Salihi, Tedesco,
  Seierstad, Kristiansen, Lang, Alexe, et~al.]{nipen2026bris}
Thomas~Nils Nipen, H{\aa}vard~Homleid Haugen, Magnus~Sikora Ingstad,
  Even~Marius Nordhagen, Aram Farhad~Shafiq Salihi, Paulina Tedesco,
  Ivar~Ambj{\o}rn Seierstad, J{\o}rn Kristiansen, Simon Lang, Mihai Alexe,
  et~al.
\newblock Regional data-driven weather modeling with a global stretched grid.
\newblock \emph{Artificial Intelligence for the Earth Systems}, 5\penalty0
  (2):\penalty0 250001, 2026.

\bibitem[Oskarsson et~al.(2023)Oskarsson, Landelius, and
  Lindsten]{oskarsson2023graph}
Joel Oskarsson, Tomas Landelius, and Fredrik Lindsten.
\newblock Graph-based neural weather prediction for limited area modeling.
\newblock \emph{arXiv preprint arXiv:2309.17370}, 2023.

\bibitem[Oskarsson et~al.(2024)Oskarsson, Landelius, Deisenroth, and
  Lindsten]{oskarsson2024probabilistic}
Joel Oskarsson, Tomas Landelius, Marc~P Deisenroth, and Fredrik Lindsten.
\newblock Probabilistic weather forecasting with hierarchical graph neural
  networks.
\newblock \emph{Advances in Neural Information Processing Systems},
  37:\penalty0 41577--41648, 2024.

\bibitem[Pathak et~al.(2022)Pathak, Subramanian, Harrington, Raja,
  Chattopadhyay, Mardani, Kurth, Hall, Li, Azizzadenesheli,
  et~al.]{pathak2022fourcastnet}
Jaideep Pathak, Shashank Subramanian, Peter Harrington, Sanjeev Raja, Ashesh
  Chattopadhyay, Morteza Mardani, Thorsten Kurth, David Hall, Zongyi Li, Kamyar
  Azizzadenesheli, et~al.
\newblock Fourcastnet: A global data-driven high-resolution weather model using
  adaptive fourier neural operators.
\newblock \emph{arXiv preprint arXiv:2202.11214}, 2022.

\bibitem[Pathak et~al.(2026)Pathak, Cohen, Garg, Harrington, Brenowitz, Durran,
  Mardani, Vahdat, Xu, Kashinath, et~al.]{pathak2026stormcast}
Jaideep Pathak, Yair Cohen, Piyush Garg, Peter Harrington, Noah Brenowitz, Dale
  Durran, Morteza Mardani, Arash Vahdat, Shaoming Xu, Karthik Kashinath, et~al.
\newblock Kilometer-scale convection-allowing model emulation using generative
  diffusion modeling.
\newblock \emph{Science Advances}, 12\penalty0 (5):\penalty0 eadv0423, 2026.

\bibitem[P{\'e}rez et~al.(2024)P{\'e}rez, Cruz, Mart{\'\i}n, and
  Guti{\'e}rrez]{perez2024transformer}
Antonio P{\'e}rez, Mario~Santa Cruz, Daniel~San Mart{\'\i}n, and
  Jos{\'e}~Manuel Guti{\'e}rrez.
\newblock Transformer based super-resolution downscaling for regional
  reanalysis: Full domain vs tiling approaches.
\newblock \emph{arXiv preprint arXiv:2410.12728}, 2024.

\bibitem[Price et~al.(2025)Price, Sanchez-Gonzalez, Alet, Andersson, El-Kadi,
  Masters, Ewalds, Stott, Mohamed, Battaglia, et~al.]{price2025gencast}
Ilan Price, Alvaro Sanchez-Gonzalez, Ferran Alet, Tom~R Andersson, Andrew
  El-Kadi, Dominic Masters, Timo Ewalds, Jacklynn Stott, Shakir Mohamed, Peter
  Battaglia, et~al.
\newblock Probabilistic weather forecasting with machine learning.
\newblock \emph{Nature}, 637\penalty0 (8044):\penalty0 84--90, 2025.

\bibitem[Ridal et~al.(2024)Ridal, Bazile, Le~Moigne, Randriamampianina,
  Schimanke, Andrae, Berggren, Brousseau, Dahlgren, Edvinsson,
  et~al.]{ridal2024cerra}
Martin Ridal, Eric Bazile, Patrick Le~Moigne, Roger Randriamampianina, Semjon
  Schimanke, Ulf Andrae, Lars Berggren, Pierre Brousseau, Per Dahlgren, Lisette
  Edvinsson, et~al.
\newblock Cerra, the copernicus european regional reanalysis system.
\newblock \emph{Quarterly Journal of the Royal Meteorological Society},
  150\penalty0 (763):\penalty0 3385--3411, 2024.

\bibitem[Schillinger et~al.(2026)Schillinger, Samarin, Shen, Knutti, and
  Meinshausen]{schillinger2026enscale}
Maybritt Schillinger, Maxim Samarin, Xinwei Shen, Reto Knutti, and Nicolai
  Meinshausen.
\newblock Enscale: Temporally-consistent multivariate generative downscaling
  via proper scoring rules.
\newblock \emph{Journal of Advances in Modeling Earth Systems}, 18\penalty0
  (8):\penalty0 e2025MS005593, 2026.

\bibitem[Schulz \& Lerch(2022)Schulz and Lerch]{schulz2022machine}
Benedikt Schulz and Sebastian Lerch.
\newblock Machine learning methods for postprocessing ensemble forecasts of
  wind gusts: A systematic comparison.
\newblock \emph{Monthly Weather Review}, 150\penalty0 (1):\penalty0 235--257,
  2022.

\bibitem[Tudor \& Termonia(2010)Tudor and Termonia]{tudor2010boundary}
Martina Tudor and Piet Termonia.
\newblock Alternative formulations for incorporating lateral boundary data into
  limited-area models.
\newblock \emph{Monthly weather review}, 138\penalty0 (7):\penalty0 2867--2882,
  2010.

\bibitem[Vaswani et~al.(2017)Vaswani, Shazeer, Parmar, Uszkoreit, Jones, Gomez,
  Kaiser, and Polosukhin]{vaswani2017attention}
Ashish Vaswani, Noam Shazeer, Niki Parmar, Jakob Uszkoreit, Llion Jones,
  Aidan~N Gomez, {\L}ukasz Kaiser, and Illia Polosukhin.
\newblock Attention is all you need.
\newblock \emph{Advances in neural information processing systems}, 30, 2017.

\bibitem[Warner et~al.(1997)Warner, Peterson, and Treadon]{warner1997lateral}
Thomas~T Warner, Ralph~A Peterson, and Russell~E Treadon.
\newblock A tutorial on lateral boundary conditions as a basic and potentially
  serious limitation to regional numerical weather prediction.
\newblock \emph{Bulletin of the American Meteorological Society}, 78\penalty0
  (11):\penalty0 2599--2618, 1997.

\bibitem[Xu et~al.(2026)Xu, Sumbul, Mi, Zbinden, and Tuia]{xu2026geofar}
Chang Xu, Gencer Sumbul, Li~Mi, Robin Zbinden, and Devis Tuia.
\newblock Geofar: Geography-informed frequency-aware super-resolution for
  climate data.
\newblock In \emph{The Fourteenth International Conference on Learning
  Representations}, 2026.

\bibitem[Xu et~al.(2025)Xu, Zheng, Gao, Wang, Yin, Zhang, Zhang, Luo, Wang,
  Zhang, et~al.]{xu2025yinglong}
Pengbo Xu, Xiaogu Zheng, Tianyan Gao, Yu~Wang, Junping Yin, Juan Zhang, Xuanze
  Zhang, San Luo, Zhonglei Wang, Zhimin Zhang, et~al.
\newblock An artificial intelligence-based limited area model for forecasting
  of surface meteorological variables.
\newblock \emph{Communications Earth \& Environment}, 6\penalty0 (1):\penalty0
  372, 2025.

\end{thebibliography}
\end{document}